\documentclass[conference]{IEEEtran}

\usepackage{cite}
\usepackage{amsmath, amssymb, amsfonts}
\usepackage{graphicx}
\usepackage{textcomp}
\usepackage{multirow}
\usepackage{array}
\usepackage{svg}
\usepackage{booktabs}
\usepackage{rotating}
\usepackage{hyperref}
\usepackage{fancyhdr}
\usepackage{lmodern}
\usepackage{adjustbox}
\usepackage{ragged2e}
\usepackage{balance}
\usepackage{amssymb}  
\usepackage{pifont} 
\usepackage{makecell}

\hypersetup{
    colorlinks=true,
    urlcolor=blue,
    citecolor=black,
    pdftitle={Overleaf Example},
    pdfpagemode=FullScreen,
}
\usepackage[utf8]{inputenc}  
\usepackage{textcomp}        

\DeclareUnicodeCharacter{22C6}{\textasteriskcentered}
\DeclareUnicodeCharacter{2212}{\ensuremath{-}}

\author{
\IEEEauthorblockN{
    Tanisha Singh\IEEEauthorrefmark{1}, 
    Shreshtha Jha\IEEEauthorrefmark{1}, 
    Nidhi Bhatt\IEEEauthorrefmark{2}, 
    Palak Handa\IEEEauthorrefmark{3}, \\
    Nidhi Goel\IEEEauthorrefmark{1}, 
    S. Indu\IEEEauthorrefmark{4}
}
\vspace{5mm}
\IEEEauthorblockA{\IEEEauthorrefmark{1}Department of Electronics and Communication Engineering, 
Indira Gandhi Delhi Technical \\University for Women, Delhi, India}
\IEEEauthorblockA{\IEEEauthorrefmark{2}Department of Computer Science Engineering, Indira Gandhi Delhi Technical \\University for Women, Delhi, India}
\IEEEauthorblockA{\IEEEauthorrefmark{3}Research Center for Medical Image Analysis and Artificial Intelligence, 
Department of Medicine, \\Danube Private University, Krems, Austria}
\IEEEauthorblockA{\IEEEauthorrefmark{4}Department of Electronics and Communication Engineering, Delhi Technological University, Delhi, India}
}

\title{Artificial Intelligence in Gastrointestinal Bleeding Analysis for Video Capsule Endoscopy: Insights, Innovations, and Prospects (2008-2023)}

\begin{document}

\maketitle

\begin{abstract}
\textit{}{The escalating global mortality and morbidity rates associated with gastrointestinal (GI) bleeding, compounded by the complexities and limitations of traditional endoscopic methods, underscore the urgent need for a critical review of current methodologies used for addressing this condition. With an estimated 300,000 annual deaths worldwide, the demand for innovative diagnostic and therapeutic strategies is paramount. The introduction of Video Capsule Endoscopy (VCE) has marked a significant advancement, offering a comprehensive, non-invasive visualization of the digestive tract that is pivotal for detecting bleeding sources unattainable by traditional methods. Despite its benefits, the efficacy of VCE is hindered by diagnostic challenges, including time-consuming analysis and susceptibility to human error. This backdrop sets the stage for exploring Machine Learning (ML) applications in automating GI bleeding detection within capsule endoscopy, aiming to enhance diagnostic accuracy, reduce manual labor, and improve patient outcomes. Through an exhaustive analysis of 113 papers published between 2008 and 2023, this review assesses the current state of ML methodologies in bleeding detection, highlighting their effectiveness, challenges, and prospective directions. It contributes an in-depth examination of AI techniques in VCE frame analysis, offering insights into open-source datasets, mathematical performance metrics, and technique categorization. The paper sets a foundation for future research to overcome existing challenges, advancing gastrointestinal diagnostics through interdisciplinary collaboration and innovation in ML applications.}
\end{abstract}


\section{Introduction}\label{sec1}
\label{intro}

Gastrointestinal (GI) bleeding, a critical medical condition, involves the hemorrhage within the digestive tract encompassing the esophagus, stomach, small intestine, large intestine (colon), rectum, and anus. This pathological state presents a spectrum of risks, from immediate life-threatening emergencies to long-term detrimental consequences \cite{Lewis2002, Lee2004}. The severity of GI bleeding ranges from asymptomatic microhemorrhages detectable only by laboratory tests to significant visible bleeding, manifesting in stool or vomit, potentially culminating in fatal outcomes. The emergence of GI bleeding stems from various underlying conditions, including peptic ulcers, diverticulosis, inflammatory bowel disease (IBD), and colorectal cancer, each contributing to the complexity and danger of this condition. The World Health Organization (WHO) underscores the gravity of this issue, attributing approximately 300,000 annual deaths worldwide to GI bleeding \cite{Wilcox2009}. These daunting statistics underscore the need for continued research and innovation in diagnostic and therapeutic approaches.

GI bleeding is broadly classified into upper and lower GI bleeding, each with distinct characteristics and implications \cite{DiGregorio2023}. Upper GI bleeding, more prevalent than its lower counterpart primarily originates from the upper digestive tract \cite{Wuerth2018}. The incidence of upper GI bleeding ranges from 50 to 150 cases per 100,000 population annually \cite{Salyer2007}, disproportionately affecting older individuals \cite{Rockey2005}. Mortality rates for upper GI bleeding lie between 6\% to 10\%, reflecting its significant health impact \cite{Vreeburg1997, Yavorski1995, WilcoxClark1999}. Conversely, lower GI bleeding, with an annual incidence of 20-27 episodes per 100,000 persons, has a slightly lower mortality rate of 4\%-10\% \cite{Hussain2000, Zuccaro1998}. Acute GI bleeding, a significant cause of hospital admissions in the United States, is estimated at 300,000 patients annually \cite{Rubin, ManningDimmitt2005}. The clinical presentation of GI bleeding can be overt, such as hematemesis, melena, hematochezia, or occult, often detected through positive fecal occult blood tests or iron deficiency anemia \cite{BritishSociety2002}. The complexity extends to obscure GI bleeding, where the bleeding source remains unidentified after standard endoscopic evaluations \cite{BullHenry2013, Raju2007}.

Diagnostic modalities for GI bleeding have evolved, with endoscopy being the primary method for upper GI bleeding \cite{Adang1995}. It involves direct visualization of the GI tract to identify bleeding sources. However, the advent of Video Capsule Endoscopy (VCE) marks a significant advancement. This method has shown a higher diagnostic yield in patients with chronic occult and obscure GI bleeding compared to push enteroscopy and barium studies \cite{Triester2005}. VCE, a non-invasive and painless procedure, involves swallowing a small, wireless capsule containing a camera that captures frames as it travels through the digestive tract. It provides a comprehensive visualization of the small intestine, which is not easily accessible by traditional endoscopy methods. It offers the advantage of minimal or no hospitalization and specialized care. It does not require sedation, and the capsule is naturally excreted from the body. It plays a crucial role in identifying the source and location of bleeding. Thus, it can help in detecting lesions, ulcers, tumors, abnormal blood vessels, and other abnormalities \cite{Iddan2000}. In developed countries, VCE is being utilized as the first line of treatment for the management of GI conditions. However, the process of developing a differential diagnosis based on the 6-8 hours of video captured is time-consuming. The evaluation of anomalies in a frame-by-frame manner requires expertise and is susceptible to human error. Moreover, due to the movement of the camera along the GI tract, there are challenges such as poor lighting, motion artifacts, blurring, and the possibility of missed abnormalities at the camera's rear side \cite{khan2020}. Hence, automation in VCE can streamline the process by providing standardized and objective analysis. It saves the time of healthcare professionals, enabling faster diagnosis and treatment. Furthermore, it enhances detection accuracy and minimizes the rate of false positives or false negatives \cite{Kim2021}.

\begin{figure}[htbp]
    \centering
    \includegraphics[width=\columnwidth]{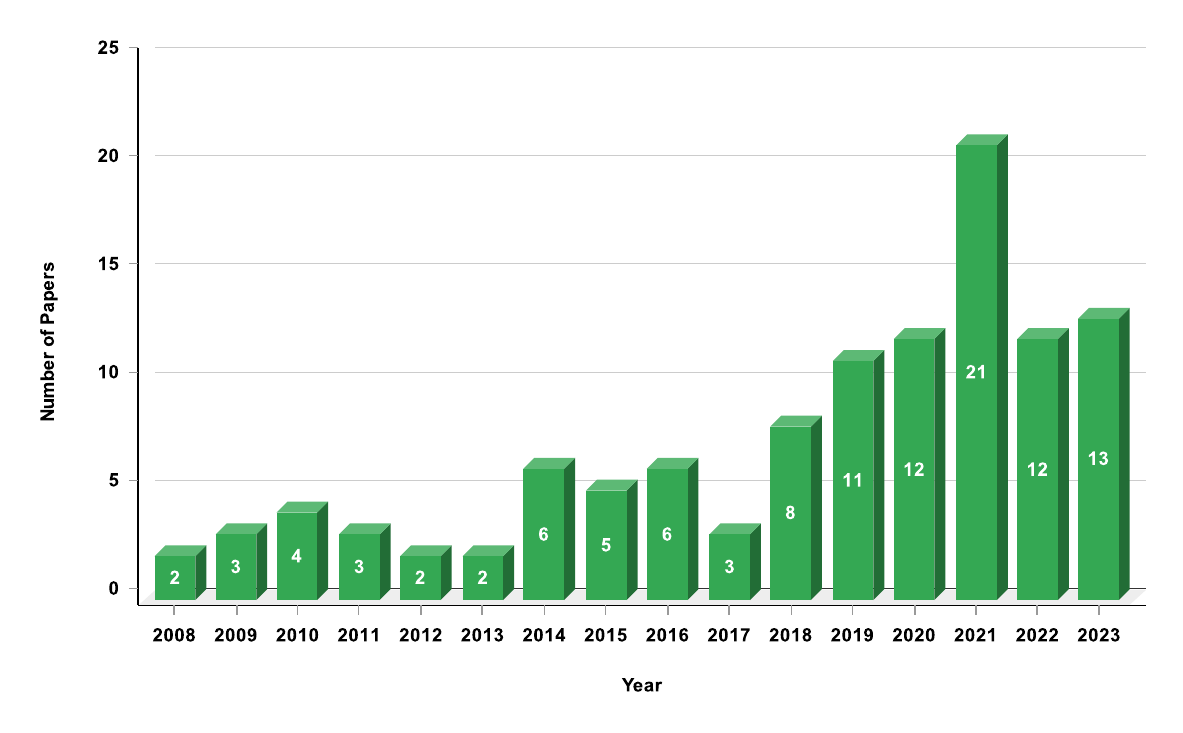}
    \caption{Year-wise distribution of papers included in the review.}
    \label{fig:yearwise_distribution}
\end{figure}

These challenges have catalyzed interest in employing Machine Learning (ML) for the automatic classification, detection, and segmentation of GI bleeding in VCE. ML advancements promise improved accuracy, faster diagnostic processes, and reduced human error. This review paper examines the current ML methodologies for automatic bleeding detection in VCE, assessing their effectiveness, challenges, and future directions. It analyzes 113 papers published between 2008 and 2023, with Figure \ref{fig:yearwise_distribution} depicting the year-wise distribution of these studies, highlighting the growing interest and research efforts in this field.

\subsection{Motivation}\label{subsec1}

The literature indicates a rising incidence of GI bleeding and the inadequacies of conventional diagnostic techniques, highlighting the urgent demand for novel solutions that employ advanced methods to increase diagnostic precision, decrease manual effort, and enhance patient outcomes. The following key factors influence the drive to pursue this comprehensive review:

\begin{itemize}
\item \textbf{Advancements in Deep Learning (DL) and ML}: Recent breakthroughs in DL have demonstrated remarkable success in medical image analysis. The exploration of the application of these advanced methodologies in this domain is thus imperative.

\item \textbf{Evaluation Parameters}: Diverse evaluation parameters play a critical role in validating the effectiveness of GI bleeding detection methodologies. By exploring these parameters, we aim to identify the most relevant metrics that ensure the reliability and robustness of diagnostic techniques.

\item \textbf{Exploration of Open-Source Datasets}: The availability of open-source data for GI bleeding offers a foundation for developing and testing AI models. We intend to review these datasets for the development and validation of more accurate, universally applicable ML models.

\item \textbf{Technique Categorization and Progress Discussion}: A systematic categorization of the techniques used in the detection, classification, and segmentation of GI bleeding is essential. By examining the progress and advancements in each area, we aim to identify trends, gaps, and opportunities for future research in improving GI bleeding diagnosis through ML and image processing technologies.
\end{itemize}

\subsection{Contribution}\label{subsec2}
This paper provides an exhaustive analysis of AI techniques used to detect GI bleeding in VCE frames. Our contributions are summarized as follows:

\begin{enumerate}
\item \textbf{GI Bleeding Overview}: We present an overview of GI bleeding, including its causes, symptoms, and the role of VCE in its diagnosis.

\item \textbf{Critical Literature Analysis}: Evaluates current research on ML for GI bleeding detection, identifying gaps and future opportunities.

\item \textbf{Open Source Datasets Analysis}: Provides the first extensive assessment of open source datasets for abnormal bleeding detection in VCE frames, laying the groundwork for future model development.

\item \textbf{Mathematical Performance Metrics}: Details mathematical formulas for diagnostic accuracy and precision metrics in GI bleeding detection.

\item \textbf{Technique Examination}: An in-depth examination of current techniques employed in GI bleeding detection, including a distinction between classification, segmentation, detection, and combined approaches.

\item \textbf{Future Research Directions}: Suggests potential research avenues to leverage ML advancements for better GI bleeding detection. Identification of gaps in current research, proposing possible future research directions.

\end{enumerate}

\begin{figure}[htbp]
  \centering
  \includegraphics[width=\columnwidth]{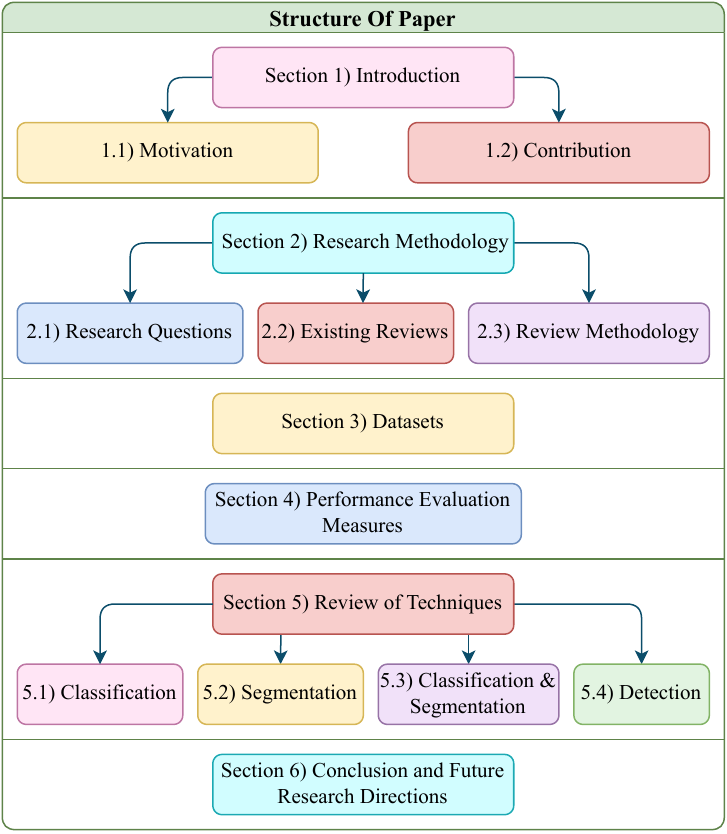} 
  \caption{Flowchart depicting the structure of the review.}
  \label{fig:structure_of_paper}
\end{figure}

To the best of our knowledge, this is the first review to provide such an in-depth analysis explicitly focused on the application of ML for GI bleeding detection in VCE. Through this review, we aim to offer insights that could guide future researchers, provide foundational knowledge, and stimulate advancements in GI healthcare research. Fig. \ref{fig:structure_of_paper} outlines our review structure: Section \ref{sec1} introduces the motivation and contribution of the study. Section \ref{Sec2} details the research methodology, including research questions, existing reviews, and the review approach. Section \ref{sec333} discusses the open source datasets for abnormal bleeding detection in VCE frames. Performance evaluation measures are covered in Section \ref{sec4}, divided into statistical, classification, and segmentation metrics. Section \ref{Sec5} reviews the techniques used in abnormal bleeding detection, including classification, segmentation, and combined approaches. The paper concludes with Section \ref{sec6}, presenting concluding remarks and future research directions.

\section{Research Methodology}\label{Sec2}
This section outlines the methodology for analyzing traditional and ML techniques in detecting GI bleeding in VCE frames, covering research questions, existing reviews, and the selection process for literature review.

\begin{table}[htbp]
\centering
\caption{Research questions related to GI bleeding detection in VCE frames.}
\small
\begin{tabular}{|c|p{6cm}|}
\hline
\textbf{S. No.} & \textbf{Research Questions} \\
\hline
Q1 & What is video capsule endoscopy, and how is it used in bleeding detection? \\
\hline
Q2 & What are the primary types of bleeding detectable via VCE? \\
\hline
Q3 & What are the various ML techniques applied in VCE bleeding detection? \\
\hline
Q4 & What datasets are most commonly used in ML models for VCE bleeding detection? \\
\hline
Q5 & What are the key performance evaluation metrics for ML models in this field? \\
\hline
Q6 & What are the challenges in detecting bleeding using ML in VCE? \\
\hline
Q7 & How do transparency and interpretability in DL models impact VCE bleeding detection? \\
\hline
Q8 & How does ML in VCE bleeding detection compare to traditional image processing and non-AI methods? \\
\hline
Q9 & What are the future research directions and unexplored areas in ML for VCE bleeding detection? \\
\hline
\end{tabular}
\vspace{-0.5\baselineskip} 
\label{table:research-questions}
\end{table}


\subsection{Research Questions}\label{subsec3_rq}

This review aims to be a definitive guide for researchers exploring the intersection of VCE and ML in bleeding detection. It presents a series of research questions aimed at unpacking the nuances of this specialized domain. These questions not only illuminate key aspects of VCE bleeding detection but also highlight open challenges and future directions for research. Table \ref{table:research-questions} outlines these research questions, providing a structured path for emerging researchers to understand the foundational and advanced aspects of this field. Meanwhile, Table \ref{table:comparison_intermsof_research_questions} offers a comparative analysis of existing literature, showing how each survey, including ours, addresses these questions. This comparison highlights the unique and comprehensive nature of our review and its contribution to the field's knowledge base.

\begin{table*}[htbp]
\centering
\caption{Comparative analysis of review papers on bleeding detection in VCE frames in terms of research questions.} \scriptsize
\begin{adjustbox}
{width=\textwidth,totalheight=\textheight,keepaspectratio}
\renewcommand{\arraystretch}{1.2}
\setlength{\tabcolsep}{4pt}
\begin{tabular}{|>{\RaggedRight\hspace{0pt}}p{2.7cm}|>{\RaggedRight\hspace{0pt}}p{1.1cm}|>{\RaggedRight\hspace{0pt}}p{0.8cm}|>{\RaggedRight\hspace{0pt}}p{0.8cm}|>{\RaggedRight\hspace{0pt}}p{0.8cm}|>{\RaggedRight\hspace{0pt}}p{0.8cm}|>
{\RaggedRight\hspace{0pt}}p{0.8cm}|>
{\RaggedRight\hspace{0pt}}p{0.8cm}|>
{\RaggedRight\hspace{0pt}}p{0.8cm}|>
{\RaggedRight\hspace{0pt}}p{0.8cm}|>{\RaggedRight\hspace{0pt}}p{0.8cm}|}
\hline
    \textbf{Study} & \textbf{Year} & \textbf{Q1} & \textbf{Q2} & \textbf{Q3} & \textbf{Q4} & \textbf{Q5} & \textbf{Q6} & \textbf{Q7} & \textbf{Q8} & \textbf{Q9} \\
    \hline
    Brzeski et al. & 2013 & \checkmark & \ding{55} & \checkmark & \ding{55} & \ding{55} & \checkmark & \ding{55} & \ding{55} & \checkmark \\
    \hline
    Du et al. & 2019 & \checkmark & \ding{55} & \checkmark & \ding{55} & \ding{55} & \checkmark & \ding{55} & \checkmark & \checkmark \\
    \hline
    Rahim et al. & 2020 & \checkmark & \ding{55} & \checkmark & \checkmark & \ding{55} & \ding{55} & \ding{55} & \ding{55} & \checkmark \\
    \hline
    Soffer et al. & 2020 & \checkmark & \ding{55} & \checkmark & \checkmark & \ding{55} & \checkmark & \ding{55} & \ding{55} & \checkmark \\
    \hline
    Musha et al. & 2023 & \checkmark & \ding{55} & \checkmark & \checkmark & \ding{55} & \checkmark & \ding{55} & \ding{55} & \checkmark \\
    \hline
\end{tabular}
\end{adjustbox}
\label{table:comparison_intermsof_research_questions}
\end{table*}

\subsection{Existing Reviews}\label{subsec4}

This section details several relevant review articles aimed at summarizing techniques for detecting bleeding in VCE frames. The section also attempts to explain the limitations of these review articles by bench-marking them with our presented review. Extensive reviews have addressed the need for ML algorithms in automatic bleeding classification, detection, and segmentation in VCE in a timely manner. Increasing scientific advancements in image processing methods, ML, transfer learning (TL), and DL in bleeding classification, detection, and segmentation in VCE frames have been observed in \cite{brzeski2013survey, rahim2020survey, du2019review, soffer2020deep, musha2023review}.

Brzeski et al. (2013) \cite{brzeski2013survey} conducted a comprehensive analysis of existing methods for detecting bleeding in VCE, emphasizing image processing techniques utilized in algorithms. They categorized these techniques into various groups: Color-based, Texture-based, Contour-based, Segmentation, Decision, and Other. The authors concluded that algorithms may need broader techniques to address visual features in endoscopic bleeding. While Brzeski et al.'s analysis provides valuable insights into image processing methodologies, their review lacks coverage of DL techniques specifically applied to bleeding detection in VCE. Additionally, there is no discussion on datasets commonly used in DL models for bleeding detection or the comparative analysis between DL and traditional image processing methods in this context. Du et al. (2019) \cite{du2019review} provided an extensive review of the application of DL methods in the analysis of GI frames. The paper outlined various DL architectures, including Convolutional Neural Networks (CNN), supervised architectures like LeNet, AlexNet, Visual Geometry Group Network (VGGNet), GoogleNet, Residual Network (ResNet), unsupervised architectures such as Generative Adversarial Networks (GAN), and others. The review covered existing works on classification, detection, and segmentation related to GI diseases, including polyps, hemorrhages, GI cancer, and GI disease analysis. The findings highlighted polyps as the most extensively studied GI disease, with GI cancer being the second most explored. The authors found that CNN-based supervised learning networks currently dominate the landscape of deep models in GI frame processing. The authors suggested areas for further exploration, including developing a 3D-CNN-based DL diagnostic system and exploring other DL methods, such as Recurrent Neural Networks (RNNs) and Graph Neural Networks (GNNs). Although Du et al.'s review offers a detailed overview of DL methods in GI frame analysis, it lacks a specific focus on bleeding detection in VCE. 

Rahim et al. (2020) \cite{rahim2020survey} conducted a review focusing on computer-aided detection (CAD) methods utilizing VCE frames. The primary objective was to classify these frames as either diseased/abnormal or disease-free/normal, specifically targeting bleeding, tumors, polyps, and ulcers. The survey encompasses a comprehensive analysis of various studies, providing detailed insights into their methodologies, findings, and conclusions. The authors noted that no existing research has addressed the joint classification of all three anomalies. Thus, they proposed an innovative approach for the concurrent classification of tumors, polyps, and ulcers in VCE videos. This proposed method employs a cascaded approach of neural networks, and the paper includes pertinent information on the experimental setup, the dataset used, and the obtained results. While Rahim et al.'s review provides insights into CAD methods in VCE, it lacks coverage of DL techniques and their application, specifically in bleeding detection. Additionally, there is no discussion on the comparative analysis between DL and traditional image processing methods in bleeding detection or future research directions in this field. Soffer et al. (2020) \cite{soffer2020deep} presented the findings of a systematic review of the current literature for DL implementation in VCE. The work provides a succinct overview of the characteristics of articles included in a quantitative meta-analysis, specifically focusing on clinical applications, dataset sizes, and network performance, particularly on ulcers and bleeding. By examining existing research, CNN is concluded to show remarkable performance in abnormality detection, with accuracy consistently exceeding 90\% in most studies. Quantitative analysis showed high pooled sensitivities and specificities for the detection of ulcers, bleeding, or the detection of the source of bleeding. Although the systematic review by Soffer et al. offers valuable insight into DL implementation in VCE, it lacks a detailed discussion of bleeding detection techniques, challenges, and future research directions specific to this application. Additionally, there is no coverage of datasets commonly used in DL models for bleeding detection or comparative analysis with traditional image processing methods.

Musha et al. (2023) \cite{musha2023review} conducted a systematic review of CAD bleeding algorithms in VCE. The study aimed to identify a taxonomy for these algorithms, discuss prevalent methods, and pinpoint the most effective ones for practical use. The researchers categorized the existing literature into three tasks, i.e., classification, segmentation, and combined classification and segmentation. The combined task proved to be the most effective, with the highest accuracy and specificity. In the review, various color spaces were divided into four groups, i.e., RGB (red, green, blue), HSV (hue, saturation, value), combined (multiple color spaces), and other (YIQ (Luminance (Y), In-phase (I), Quadrature (Q)), YCbCr (Luma (Y), Blue Chroma (Cb), Red Chroma (Cr)), etc.). As per the review above, the selected color space did not offer any performance advantages for the classification or segmentation tasks except RGB, which showed slightly higher recall and lower variance. Combining global and local feature extraction yielded better accuracy. The research findings reveal that for ML, K-Nearest Neighbour (KNN), Support Vector Machine (SVM), Multi-Layer Perceptron (MLP), and neural network algorithms were commonly used, while CNNs dominated DL. Although KNN and CNN outperformed other algorithms, KNN faced limitations like overfitting and reliance on hand-crafted features. 

In the above review, the chosen color space did not provide any performance benefits for the classification or segmentation tasks. The study highlighted the growing focus on DL algorithms, particularly CNNs, for bleeding detection and emphasized the need to address dataset limitations by using synthetic VCE frames. Challenges include minimizing computational requirements and ensuring visually interpretable outcomes for clinician trust. While Musha et al.'s systematic review provides valuable insights into bleeding detection algorithms in VCE, it lacks coverage of DL techniques and their application specifically in VCE. Additionally, there is no discussion on datasets commonly used in DL models for bleeding detection or comparative analysis with traditional image processing methods. By addressing these gaps, our review offers a comprehensive understanding of bleeding detection in VCE, covering all research questions listed in Table \ref{table:research-questions} and providing valuable insights into DL techniques, datasets, challenges, and future research directions in this field.

\subsection{Review Approach}\label{subsec5_review}

\begin{figure}[htbp]
  \centering
  \includegraphics[width=\columnwidth]{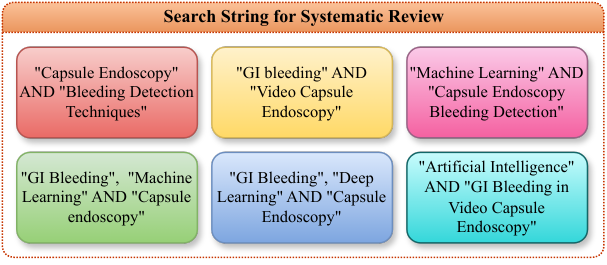} 
  \caption{Search string for searching the research articles.}
  \label{fig:Search_string}
\end{figure}

This review on GI bleeding detection follows the Preferred Reporting Items for Systematic Reviews
and Meta-Analyses (PRISMA) guidelines \cite{page2021prisma}. PRISMA is selected for its robust framework, which helps outline the processes for content retrieval and assessment to enhance the credibility and thoroughness of the review. To ensure comprehensive coverage, we utilized four academic databases: Google Scholar, PubMed, Researchgate, and IEEE Xplore. Additionally, preprint platforms such as ArXiv, TechRxiv, MedRxiv, and BioRxiv were explored to include the most recent studies. The search terms included ``Capsule Endoscopy" AND ``Bleeding Detection Techniques", ``GI bleeding" AND ``Video Capsule Endoscopy", ``Deep Learning" AND ``WCE Bleeding Detection", ``Machine Learning" AND ``GI Bleeding Detection", and ``Artificial Intelligence" AND ``GI Bleeding in
Video Capsule Endoscopy". Fig. \ref{fig:Search_string} illustrates the detailed search string used.

A rigorous, multi-phase screening process was implemented to evaluate both traditional and ML techniques for VCE bleeding detection. Initially, 500 manuscripts were identified through both electronic databases (N=400) and manual searches (N=100). After removing duplicates, 300 manuscripts remained. After reviewing titles, abstracts, and introductions, these were further narrowed down to 150 manuscripts. Ultimately, 113 manuscripts were selected based on predefined selection criteria. Selected papers ranged from 2008 to 2023, covering various methodologies, datasets, and findings. The selection process ensured that only papers with the most impacted and innovative approaches were included.

\begin{figure}[htbp]
  \centering
  \includegraphics[width=\columnwidth]{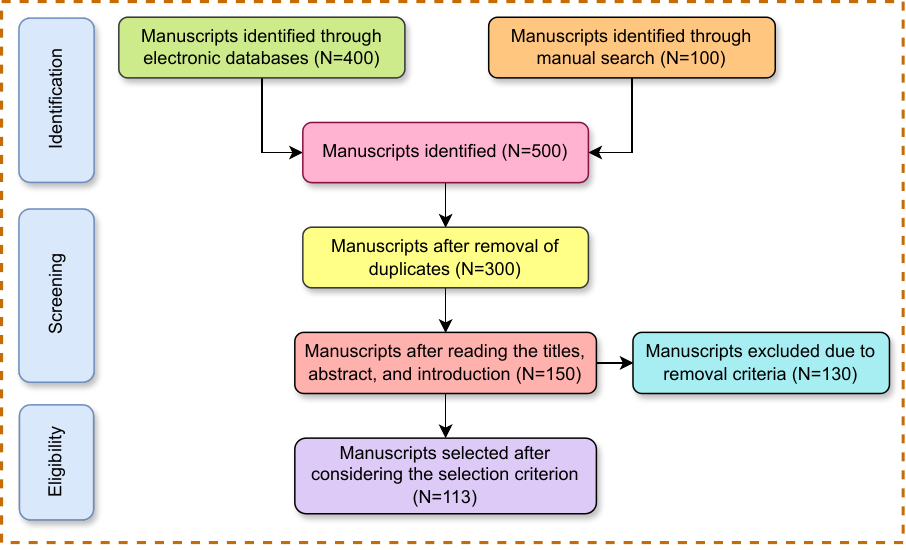} 
  \caption{PRISMA: Outlining the review strategy for bleeding detection in VCE frames.}
  \label{fig: PRISMA}
\end{figure}

\begin{table*}
\centering
\caption{Selection and removal criteria for the systematic review of VCE GI bleeding detection.}
\label{table: selection_removal}
\begin{adjustbox}{width=\textwidth}
\renewcommand{\arraystretch}{1.2}
\setlength{\tabcolsep}{4pt}
\begin{tabular}{|>{\RaggedRight}p{0.07\textwidth}|>{\RaggedRight}p{0.15\textwidth}|>{\RaggedRight}p{0.35\textwidth}|>{\RaggedRight}p{0.33\textwidth}|}
\hline
\textbf{S. No.} & \textbf{Parameter} & \textbf{Selection Criteria} & \textbf{Removal Criteria} \\ \hline
1 & Time Frame & Studies spanning from 2008 to 2023 & Studies published before 2008 \\ \hline
2 & Language & Papers written in English & Papers in languages other than English \\ \hline
3 & Study Design & Quantitative studies with empirical data & Qualitative studies or theoretical papers \\ \hline
4 & Database Access & Papers accessible through databases like Google Scholar, PubMed, ResearchGate, IEEE Xplore & Papers not accessible through specified databases \\ \hline
5 & Relevance & Studies focused on VCE GI bleeding detection & Studies not focused on VCE GI bleeding detection \\ \hline
6 & Publication Type & Papers published in journals & Papers published in conferences \\ \hline
\end{tabular}
\end{adjustbox}
\end{table*}

\small
Fig. \ref{fig: PRISMA} illustrates the PRISMA flow diagram, depicting the different phases of literature selection and the number of articles screened, assessed for eligibility, and included in the review. The selection and removal criteria for research articles are mentioned in Table \ref{table: selection_removal}. This methodology ensured a thorough and unbiased collection of literature, offering a clear understanding of advancements in GI bleeding detection techniques.
\normalsize

\section{Open Source Datasets for Abnormal Bleeding Detection in VCE frames}\label{sec333}

\begin{figure}[htbp]
  \centering
  \resizebox{\columnwidth}{!}{\includegraphics{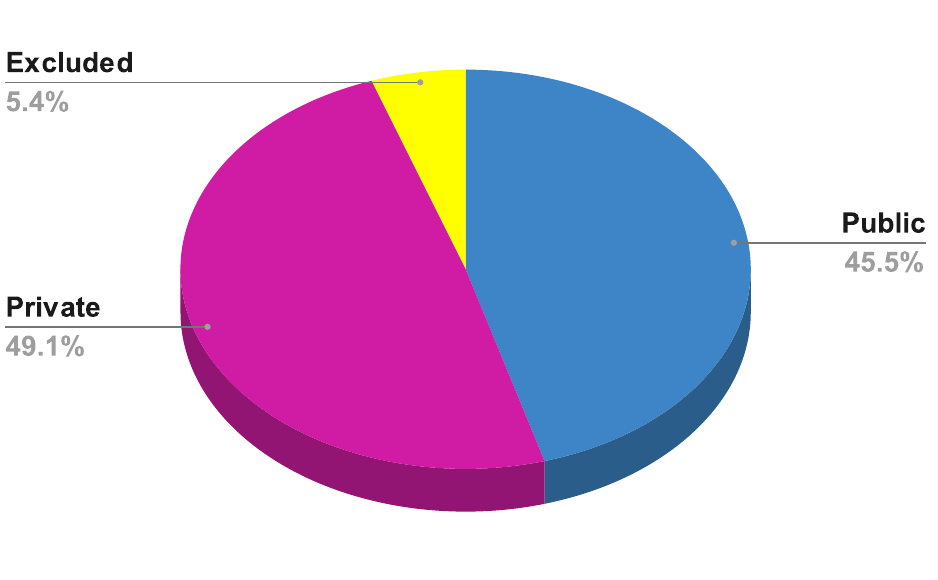}} 
  \caption{Distribution of public and private datasets used in the papers.}
  \label{fig:public_private_distribution}
\end{figure}

In the field of VCE, datasets play a significant role. The development and validation of DL models require a diverse set of datasets with varied characteristics. Researchers in the field of VCE bleeding detection employ various datasets, each with unique features \cite{singh2023automatic}. While some of these datasets are publicly accessible, others are restricted due to privacy and proprietary concerns. Fig. \ref{fig:public_private_distribution} depicts the composition of papers studied in this review that have used publicly available datasets and private datasets. There are also some papers whose information regarding the datasets used could not be found; they are labeled as excluded. Open Source datasets contribute to collaborative advancement in medical research while maintaining transparency \cite{handa2021datasets}. This section is an overview of publicly available datasets that facilitate the recent developments in bleeding detection using VCE. 

Bleeding as an abnormality was first introduced in the KID Project \cite{koulaouzidis2017kid}. It was developed to advance the medical research support systems (MDSS) for VCE. Clinicians annotated VCE frames and videos. Six centers contributed 2500 annotated VCE frames and 47 videos, which included vascular lesions that covered angioectasias and/or bleeding. KID dataset produced semantic and graphic image annotations supported by an open-access annotation tool. Kvasir-Capsule \cite{smedsrud2021kvasir}, the largest publicly available PillCAM dataset, contains 4,741,504 frames extracted from 117 videos, out of which 47,238 frames are medically verified, and are labeled with bounding box. It contains 14 different classes. Under the luminal findings section, there are two classes: blood-fresh, for the liquid red appearance of the bleeding in the upper GI tract or small bowel, and blood-hematin, for the cases of minimal bleeding where small black stripes may be observed. But, in both these datasets, bleeding is observed as a coexisting abnormality.

Red Lesion Endoscopy (RLE) Dataset \cite{coelho2018deep} contains 2 subsets of anonymized VCE frames with annotated red lesions. The first folder has 3,295 non-sequential frames with diﬀerent red lesions, such as angioectasias, angiodysplasias, bleeding, and others, which total 1,311 frames. The second folder has 600 sequential frames, out of which 73\% of frames have red lesions, each one labeled manually as Blood/Non-Blood. All frames in both folders are 320$\times$320 pixels wide. The Bleeding dataset by Farah Deeba \cite{deeba2016automated} was released in 2016 and contains 50 bleeding frames and 50 non-bleeding frames, along with ground truth frames and annotations. The limitation of this dataset was that its size does not reflect the practical cases of generating huge numbers of frames in VCE. Endoscopy Recommendation System (ERS) Dataset \cite{ersdataset_22} is a multi-label image dataset that contains frames from flexible endoscopy, colonoscopy, and Capsule Endoscopy (CE). The frames are labeled according to Minimal Standard Terminology 3.0 (MST 3.0). This dataset comprises 104 medical labels and 19 ML application labels, divided into 5 categories: healthy, blood, quality, gastro, and colono. 
WCE-BleedGen Dataset \cite{Auto_WCEbleedgen_test2023} primarily focuses on bleeding and non-bleeding frames. It consists of multiple bleeding frames from internet resources with different types of GI bleeding. It also contains medically validated binary masks and bounding boxes in You Only Look Once (YOLO)-txt, XML, and txt formats.

\small
Auto-WCEBleedGen dataset \cite{Auto_WCEbleedgen_test2023} contains independently collected bleeding and non-bleeding frames of more than 30 patients suffering from acute, obscure, and occult GI bleeding referred to the Department of Gastroenterology and HNU, All India Institute of Medical Sciences (AIIMS), New Delhi, India. It is divided into two subsets—one containing 49 frames from 7 patients with marked and unmarked annotations, and the second has 514 frames from 23 patients with known class labels but no marked annotations. The Gastrointestinal Bleeding WCE images Dataset \cite{khan2023gastrointestinal} comprises VCE frames focused on GI bleeding, sourced from Turkey Hospital Muzaffargarh (THM) in Pakistan. Collected between January 2022 and May 2022, it contains 226 GI frames. Among these, 113 frames represent instances of GI bleeding or lesions, while the remaining 113 frames capture VCE visuals from normal patients. Auto-WCEBleedGen and GI Bleeding WCE frames datasets have not been used in any manuscripts, but they focus on GI bleeding as a primary abnormality. Table \ref{table:Existing_datasets} provides a detailed description of these public datasets.
\normalsize

\begin{table*}[htbp]
\centering
\caption{Overview of Open Source VCE datasets focused on GI bleeding abnormality up to 2023.}
\label{table:Existing_datasets}
\scriptsize
\begin{adjustbox}
{width=\textwidth,totalheight=\textheight,keepaspectratio}
\renewcommand{\arraystretch}{1.2}
\setlength{\tabcolsep}{4pt}
\begin{tabular}{|>{\RaggedRight\hspace{0pt}}p{2.7cm}|>{\RaggedRight\hspace{0pt}}p{1.1cm}|>{\RaggedRight\hspace{0pt}}p{2.1cm}|>{\RaggedRight\hspace{0pt}}p{1.7cm}|>{\RaggedRight\hspace{0pt}}p{0.8cm}|>{\RaggedRight\hspace{0pt}}p{1.6cm}|>{\RaggedRight\hspace{0pt}}p{2.7cm}|}
\hline
\textbf{Name} & \textbf{Size} & \textbf{No. of bleeding frames} & \textbf{Dimension} & \textbf{CSV\textsuperscript{*}} & \textbf{Augmented frames} & \textbf{Additional Features}\\
\hline
KID \mbox{}\cite{koulaouzidis2017kid} & 335 MB & 5 & 320$\times$320 & No & No & Class labels, annotations\\
\hline
Kvasir-Capsule \mbox{}\cite{smedsrud2021kvasir} & 89 GB & 458 & 336$\times$336 & No & No & Class labels, Bounding boxes\\
\hline
RLE Set 1 and Set 2 \mbox{}\cite{coelho2018deep} & 506 MB & 448 & 320$\times$320 and 512$\times$512 & Yes & No & Annotations \\
\hline
Farah Deeba \mbox{}\cite{deeba2016automated} & 1.04 MB & 50 & 568$\times$430 & No & No & Annotations\\
\hline
ERS \mbox{}\cite{ersdataset_22} & -- & 357 & -- & Yes & Yes & Class Labels, Annotations, masks \\
\hline
WCE-BleedGen \mbox{}\cite{Auto_WCEbleedgen_test2023} & 168 MB & 1309 & 224$\times$224 & Yes & Yes (version 2) & Class labels, annotations, and bounding box \\
\hline
Auto-WCE BleedGen Dataset \mbox{}\cite{Auto_WCEbleedgen_test2023} & 89.1 MB & 339 & 224$\times$224 & No & Yes & Class labels, annotations, and bounding box\\
\hline
Gastrointestinal bleeding WCE frames \mbox{}\cite{khan2023gastrointestinal} & 596 MB & 113 & variable & No & -- & No\\
\hline
\end{tabular}
\end{adjustbox}
\end{table*}

\section{Performance Evaluation Measures}\label{sec4}

In the realm of GI bleeding detection through traditional and ML techniques, a variety of evaluation metrics are employed to assess the performance and reliability of the models. These metrics are crucial for understanding the efficacy of the models in accurately identifying bleeding instances. This section elaborates on the key metrics used in the literature.

\subsection{Statistical Metrics}\label{subsec5_sm}

\subsubsection{Mean}\label{subsec5}
The mean, commonly known as the average, is a measure of central tendency in a dataset and is calculated by summing all numerical values and dividing this sum by the count of values \cite{meanvariancekurtosis}. The mean's range can vary widely depending on the dataset. In the context of medical imaging, such as VCE, the mean is used to determine the average pixel intensity across a set of frames, aiding in tasks like normalization or contrast adjustment in the pre-processing stages of model development.  It is represented by Eq. ~\ref{eq:1}.

\begin{equation}
\label{eq:1}
\text{Mean} = \frac{\sum_{i=1}^{n} X_i}{n}  
\end{equation}

\subsubsection{Variance}\label{subsec6}
Variance is a statistical measurement of the spread between numbers in a dataset, where a high variance indicates a large spread of data points around the mean, and a low variance indicates a tight clustering of data points \cite{meanvariancekurtosis}. In VCE, variance can indicate the diversity in the dataset, such as variations in lighting, texture, and color, which can significantly impact the model's learning process. The variance is calculated using Eq. ~\ref{eq:2}.
\begin{equation}
\label{eq:2}
\text{Variance} = \frac{\sum_{i=1}^{n} (X_i - \text{Mean})^2}{n}
\end{equation}

\subsubsection{Kurtosis}\label{subsec7}
Kurtosis is a measure of the "tailedness" of the probability distribution of a dataset \cite{meanvariancekurtosis}. High kurtosis means that the data have heavy tails or outliers; low kurtosis means that the data are light-tailed or lack outliers. It is given by Eq. ~\ref{eq:3}.

\begin{equation}
\label{eq:3}
\begin{split}
\text{Kurtosis} = & \ \frac{n(n + 1)}{(n - 1)(n - 2)(n - 3)} \times \sum_{i=1}^{n} \left( \frac{X_i - \text{Mean}}{SD} \right)^4 \\
& - \frac{3(n - 1)^2}{(n - 2)(n - 3)}
\end{split}
\end{equation}

\subsubsection{Standard Deviation}\label{subsec8}
Standard deviation quantifies the amount of variation or dispersion in a set of data points, with a high standard deviation indicating that the data points are spread over a larger range of values \cite{sdv}. Its range can be from zero to positive infinity. In the realm of VCE, the standard deviation helps in understanding the variability in image features, which may affect the consistency and reliability of bleeding detection models. The formula for the standard deviation is given by Eq. ~\ref{eq:4}.

\begin{equation}
\label{eq:4}
\text{SD} = \sqrt{\frac{\sum_{i=1}^{n} (X_i - \text{Mean})^2}{n}}
\end{equation}


\subsection{Classification Metrics}\label{subsec9}

\subsubsection{Accuracy }\label{subsec10}
Accuracy is a fundamental metric representing the overall correctness of the model. It is the ratio of correctly predicted observations to the total observations \cite{accuracyprecision} and is given by Eq. ~\ref{eq:5}.

\begin{equation}
\label{eq:5}
\text{Accuracy} = \frac{\text{TP} + \text{TN}}{\text{TP} + \text{TN} + \text{FP} + \text{FN}}
\end{equation}

\subsubsection{Precision }\label{subsec11}
Precision, or positive predictive value, indicates the proportion of positive identifications that were actually correct \cite{accuracyprecision}. It is crucial in medical diagnosis to minimize the burden of unnecessary treatments. The formula for precision is given by Eq. ~\ref{eq:6}.
\begin{equation}
\label{eq:6}
\text{Precision} = \frac{\text{True Positives}}{\text{True Positives} + \text{False Positives}}
\end{equation}

\subsubsection{Sensitivity(True Positive Rate, Recall)}\label{subsec12}
Sensitivity, or recall, is a measure of the proportion of actual positives correctly identified \cite{pannu2020deep}. In the context of VCE bleeding detection, it represents the model's ability to detect actual bleeding cases. It is defined as Eq. ~\ref{eq:7}.

\begin{equation}
\label{eq:7}
\text{Sensitivity} = \frac{\text{True Positives}}{\text{True Positives} + \text{False Negatives}}
\end{equation}

\subsubsection{Specificity (True Negative Rate)}\label{subsec13}
Specificity, also known as the true negative rate, measures the proportion of actual negatives that are correctly identified. It is critical in medical diagnostics to ensure that healthy cases are not falsely diagnosed as diseased \cite{aucspecificity}. The mathematical formula for specificity is given by Eq. ~\ref{eq:8}.
\begin{equation}
\label{eq:8}
\text{Specificity} = \frac{\text{True Negatives}}{\text{True Negatives} + \text{False Positives}}
\end{equation}

\subsubsection{Area Under the Curve (AUC) - ROC Curve}\label{subsec14}
The Area Under the Curve (AUC) of the Receiver Operating Characteristic (ROC) curve is a performance measurement for classification problems at various threshold settings. The ROC curve plots the True Positive Rate (Sensitivity) against the False Positive Rate (1-Specificity), and the AUC ranges from 0 to 1 \cite{aucspecificity}, with a higher value indicating better model performance. In VCE, the AUC provides a comprehensive measure of the model's ability to distinguish between bleeding and non-bleeding cases across different thresholds.

\subsubsection{False Positive Ratio (FPR)}\label{subsec15}
The False Positive Ratio is the proportion of negative cases that are incorrectly classified as positive \cite{Liaqat2018}. It is a measure of the rate at which false alarms occur. In VCE, a lower FPR is desirable as it indicates fewer healthy patients being incorrectly diagnosed with bleeding, which is essential for patient safety and trust in the diagnostic process.
\begin{equation}
\label{eq:9}
\text{FPR} = \frac{\displaystyle\text{False Positives}}{\displaystyle\text{True Negatives} + \text{False Positives}}
\end{equation}

\subsubsection{False Negative Ratio (FNR) }\label{subsec16} The False Negative Ratio is the proportion of positive cases that are wrongly classified as negative. It is critical in medical diagnostics as it represents missed diagnoses. In VCE bleeding detection \cite{Liaqat2018}, a high FNR would mean failing to identify actual bleeding cases, which could have serious implications for patient health.
\begin{equation}
\label{eq:10}
\text{FNR} = \frac{\text{False Negatives}}{\text{True Positives} + \text{False Negatives}}
\end{equation}

\subsubsection{F1 Score}\label{subsec17}
The F-Measure, also known as the F1 Score, is a harmonic mean of precision and recall \cite{f1mcc}. It is particularly useful in scenarios where an equal balance between precision and recall is desired. The F1 Score is defined as Eq. ~\ref{eq:11}.

\begin{equation}
\label{eq:11}
\text{FPR} = \frac{\text{False Positives}}{\text{True Negatives}\,+\,\text{False Positives}}
\end{equation}

\subsubsection{Matthews Correlation Coefficient (MCC)}\label{subsec18}
The Matthews Correlation Coefficient is a measure of the quality of binary classifications. It takes into account true and false positives and negatives and is generally regarded as a balanced measure that can be used even when the classes are of very different sizes. The MCC ranges from -1 to +1 \cite{f1mcc}, where +1 represents a perfect prediction, 0 no better than random prediction, and -1 indicates total disagreement between prediction and observation. The MCC is defined as Eq. ~\ref{eq:12}.
\normalsize

\begin{equation}
\label{eq:12}
\text{MCC} = \frac{\text{TP} \times \text{TN} - \text{FP} \times \text{FN}}
{\sqrt{(\text{TP}+\text{FP})(\text{TP}+\text{FN})(\text{TN}+\text{FP})(\text{TN}+\text{FN})}}
\end{equation}

where TP, TN, FP, and FN represent the numbers of true positives, true negatives, false positives, and false negatives, respectively.

\subsubsection{Negative Predictive Value (NPV) }\label{subsec19}
NPV is the ratio of true negatives to total predicted negatives. It measures the model's ability to correctly identify negatives \cite{mamun2021}. It ranges from 0 to 1, with higher values indicating better performance. The formula for NPV is given by Eq. ~\ref{eq:13}.
\begin{equation}
\label{eq:13}
\text{NPV} = \frac{\text{TN}}{\text{TN} + \text{FN}}
\end{equation}

\subsubsection{Positive Predictive Value (PPV)}\label{subsec20}
The Positive Predicted Value, also known as precision, is the proportion of positive test results that are true positives. It ranges from 0 to 1 \cite{ppv}, with higher values indicating better performance. The formula for PPV is given by Eq. ~\ref{eq:14}.
\begin{equation}
\label{eq:14}
\text{PPV} = \frac{\text{TP}}{\text{TP} + \text{FP}}
\end{equation}

\subsubsection{Cohen's Kappa Score}\label{subsec21}
Cohen's Kappa Score is a statistic used to measure inter-rater agreement for qualitative items \cite{yogapriya2021}. It considers the possibility of the agreement occurring by chance. The Kappa Score ranges from -1 (total disagreement) to 1 (perfect agreement), with 0 indicating no agreement better than chance.  The formula for Cohen's Kappa is given by Eq. ~\ref{eq:15}.
 \begin{equation}
\label{eq:15}
 \text{Kappa} = \frac{P_o - P_e}{1 - P_e}
 \end{equation}
Where \( P_o \) is the observed agreement, and \( P_e \) is the expected agreement by chance. In VCE, Cohen's Kappa can be used to evaluate the consistency of diagnosis between different clinicians or automated systems.

\subsubsection{Fréchet Inception Distance (FID)}\label{subsec22}
FID is used to measure the quality of the generated frames \cite{XiaoDetection2022}. It calculates the distance between feature vectors calculated for real and generated frames where 
\begin{equation}
\label{eq:16}
\text{FID} = ||\mu_r - \mu_g||^2 + \text{Tr}(\Sigma_r + \Sigma_g - 2(\Sigma_r\Sigma_g)^{1/2})
\end{equation}

Where \( \mu_r, \mu_g \) are the means and \( \Sigma_r, \Sigma_g \) are the covariance matrices of real and generated frames, respectively.


\subsection{Segmentation and Object Detection Metrics}\label{subsec23}

\subsubsection{Jaccard Index}\label{subsec24}
The Jaccard Index, also known as the Intersection over Union (IoU), is a widely used metric in various fields, including medical image analysis, to compare the similarity and diversity of sample sets \cite{szczpinski2012}. In the context of VCE bleeding detection, it measures how well the predicted bleeding areas overlap with the actual bleeding areas. The formula for the Jaccard Index is given by Eq. ~\ref{eq:17}.

\begin{equation}
\label{eq:17}
\text{Jaccard Index} = \frac{|A \cap B|}{|A \cup B|}
\end{equation}
\\
In this formula,
\(|A \cap B|\) represents the intersection of the two sets A and B, which in the context of medical imaging would be the area where both the b predicted bleeding region and the actual bleeding region overlap.
\(|A \cup B|\) is the union of the two sets A and B, which is the total area covered by both the predicted and actual bleeding regions combined.
The Jaccard Index ranges from 0 to 1, where 0 indicates no overlap between the predicted and actual regions (no similarity), and 1 indicates perfect overlap (complete similarity). 

\subsubsection{Dice Score (DS)}\label{subsec25}
The Dice Score, also known as the Dice Similarity Coefficient, is a measure of overlap between two samples. This statistic \cite{hajabdollahi2019} is widely used to compare the pixel-wise agreement between a predicted segmentation and its corresponding ground truth. The Dice Score is calculated by Eq. ~\ref{eq:18}.
\begin{equation}
\label{eq:18}
\text{Dice Score (DS)} = \frac{2 \times |A \cap B|}{|A| + |B|}
\end{equation}
where A and B are the predicted and actual segmentation regions, respectively. The Dice Score ranges from 0 (no overlap) to 1 (perfect overlap).

\subsubsection{Mean Intersection over Union (mIOU)}\label{subsec26}
mIOU is an extension of the IoU metric, used primarily in multi-class segmentation tasks \cite{guan2021peak}. It calculates the IoU for each class and then averages them. This provides a more comprehensive measure for models predicting multiple classes. The formula for mIOU is given by Eq. ~\ref{eq:19}.

\begin{equation}
\label{eq:19}
\text{mIOU} = \frac{1}{N} \sum_{i=1}^{N} \text{IoU}_i
\end{equation}
where \( N \) is the number of classes, and \( \text{IoU}_i \) is the IoU for the ith class.

\subsubsection{Average Precision}\label{subsec27}
Average Precision (AP) is a widely used metric in object detection and segmentation tasks, particularly in datasets where instances are annotated with bounding boxes or segmentation masks \cite{lan2019deep}. AP summarizes the precision-recall curve as the weighted mean of precision's achieved at each threshold, emphasizing the performance at different recall levels. The range of AP is from 0 to 1, with higher values indicating better model performance.

\subsubsection{Mean Overlap Coefficient }\label{subsec28}
The Mean Overlap Coefficient is a metric used to assess the similarity and overlap between two datasets or frames \cite{khan2020gastrointestinal}. The range of the Mean Overlap Coefficient is between 0 and 1, where 0 indicates no overlap and 1 signifies complete overlap or perfect agreement between the segmented results and the ground truth.
\begin{equation}
\label{eq:20}
\text{Mean Overlap Coefficient} = \frac{1}{N} \sum_{i=1}^{N} \frac{|A_i \cap B_i|}{|A_i \cup B_i|}
\end{equation}
In this formula, \( N \) is the number of observations or samples, \( A_i \) and \( B_i \) represent the two sets (or segmented regions) for the \( i \)-th observation, and \( | \cdot | \) denotes the cardinality of the set.

\section{Overview of Techniques in Papers Conducting Bleeding Analysis}\label{Sec5}
\begin{figure}[hb!]
  \centering
  \includegraphics[scale=0.13]{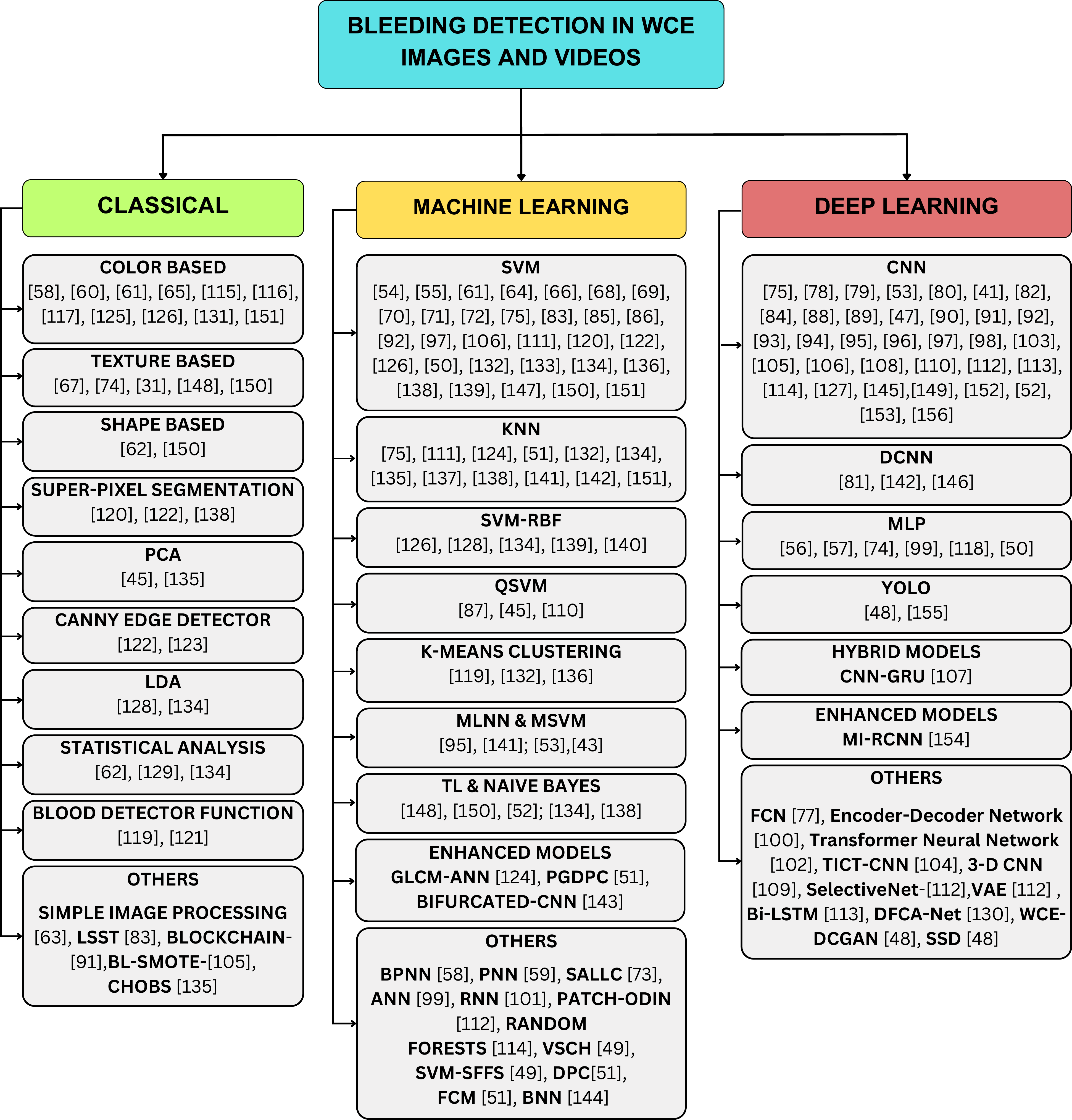}
  \vspace{1mm}
  \caption{\RaggedRight{Classifying Research Methods for Bleeding Detection in VCE frames.}}
  \label{fig:taxonomy}
\end{figure}

In this section, a comprehensive review of the techniques employed in recent studies focusing on bleeding detection in VCE frames is undertaken. Bleeding detection in VCE frames is considered a critical task in GI disease diagnosis, with applications ranging from early detection to treatment monitoring. To effectively organize the review, the reviewed articles are categorized into four distinct subsections based on their primary focus: Classification, Segmentation, Classification and Segmentation, and Detection. In the Classification subsection, papers focus solely on classifying bleeding frames in VCE frames. Segmentation papers concentrate on delineating bleeding regions within the frames. Those falling under Classification and Segmentation combine both tasks, while papers categorized under Detection primarily focus on identifying bleeding regions. The objective is to provide a detailed analysis of the novelty, methodologies, and results achieved by these papers, offering insights into the current state-of-the-art techniques in the field. This information is complemented by chord charts illustrating the relationships between authors and techniques within each subsection, as well as a comprehensive table detailing the techniques employed, datasets utilized, and the corresponding results achieved by each paper, along with the year of publication. Additionally, a taxonomy is included in the Fig. \ref{fig:taxonomy}, referencing each paper under the technique the authors have implemented in that paper. Through this review, a deeper understanding of the advancements and trends in bleeding detection methodologies within the VCE imaging domain is aimed to be facilitated.

\begin{figure}[ht]
  \centering
  \includegraphics[width=\columnwidth]{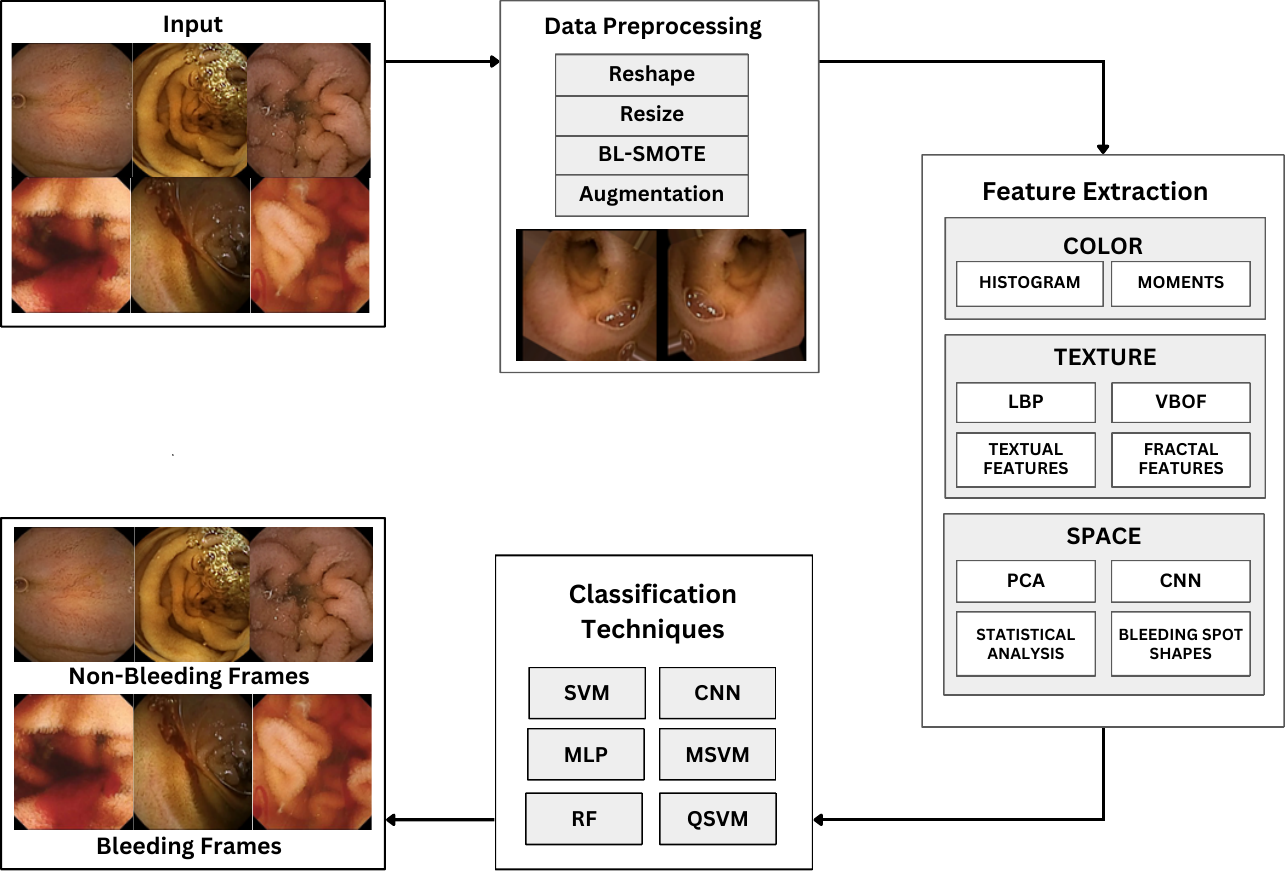}
  \vspace{1mm}
  \caption{\RaggedRight{Pipeline for Classification of Bleeding Frames in VCE frames.}}
  \label{fig:classification_pipeline}
\end{figure}

Bleeding classification in VCE frame analysis has evolved from early range ratio color condition methods to using color texture feature extraction techniques in RGB and HSI color spaces. Over time, CNNs gained prominence for feature extraction, while classifiers such as SVM were employed for classification alongside other GI anomalies like polyps, tumors, and ulcers. A recent work has integrated blockchain security with CNN for bleeding classification in VCE frames, marking a significant advancement in medical image analysis. The classification pipeline, as shown in the Fig. \ref{fig:classification_pipeline}, involves pre-processing steps such as reshaping, resizing, BorderLine Synthetic Minority Over-sampling Technique (BL-SMOTE), and data augmentation to handle imbalanced datasets. Feature extraction methods encompass color histogram, color moments, texture features (Local Binary Pattern (LBP), Visual Bag of Features (VBOF)), and space feature extraction (Principal Component Analysis (PCA), bleeding spot shapes). Metaheuristic algorithms like Binary Dragonfly Algorithm (BDA), Moth-Flame Optimization (MFO), Particle Swarm Optimization (PSO) and Crow Search Algorithm (CSA) were also employed to optimize feature extraction. Extracted features are then inputted into classifiers (SVM, MLP, Random Forest (RF) etc.) for bleeding classification. Some works employed end-to-end DL architectures like CNNs, handling the entire classification pipeline within the CNN framework.

\begin{figure}[ht]
  \centering
  \includegraphics[width=\columnwidth]{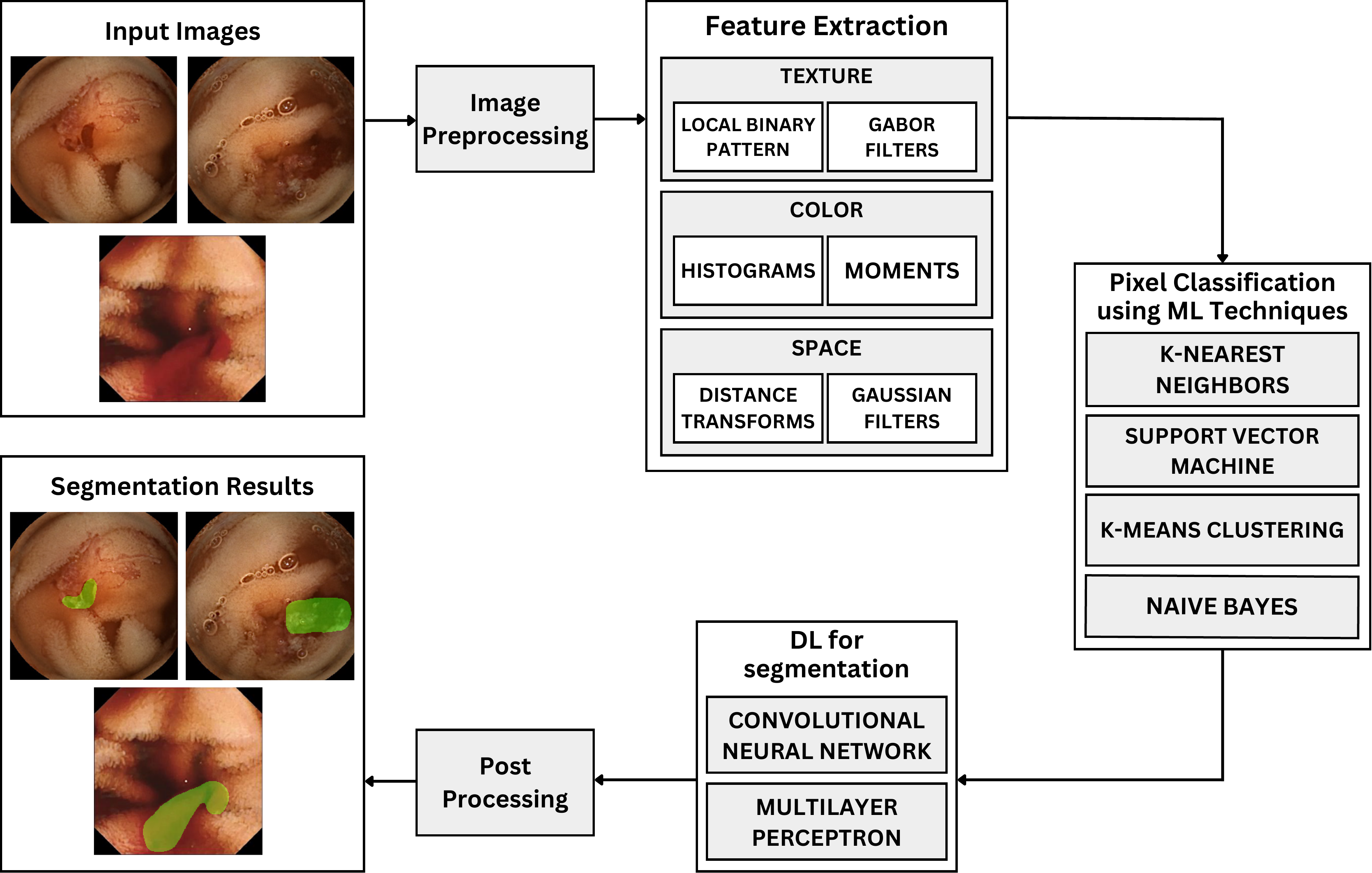} 
  \caption{\RaggedRight{Pipeline for Segmentation of Bleeding Frames in VCE images.}}
  \label{fig:segmentation_pipeline}
\end{figure}

For the purpose of segmentation of bleeding regions in VCE frames, research has developed from color and texture-based methods to advanced DL algorithms for bleeding region annotation. Some highlights in this domain include the implementation of super-pixel segmentation techniques, dual networks for accurately localizing bleeding regions, and the use of GPU-based parallel programming for reduced computational power. In the domain of detection of bleeding frames, i.e., classification, followed by segmentation of bleeding regions in VCE frames, research began by increasing algorithm proficiency using techniques like saliency maps and classifier fusion algorithms. In recent years, there has been advancement with the implementation of block-based statistics, pixel-based feature extraction, and the introduction of DL algorithms such as VCENet. After reviewing all these techniques, we have presented a generalized segmentation pipeline in Fig. \ref{fig:segmentation_pipeline} that includes pre-processing of the input image, followed by feature extraction based on color, texture, or spatial characteristics. Then, pixel-level classification of bleeding regions is applied, after which DL methods such as CNN or MLP are usually employed for accurate segmentation. This is sometimes followed by post-processing with morphological operations to enhance the segmentation results.

\begin{figure}[ht]
  \centering
  \includegraphics[width=\columnwidth]{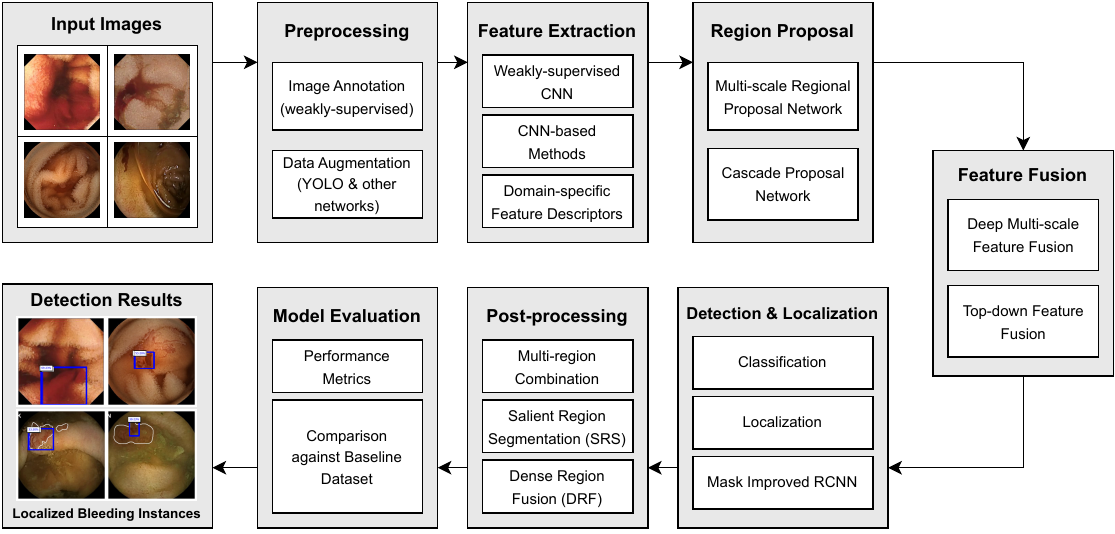}
  \caption{\RaggedRight{Pipeline for localization of Bleeding instances in VCE frames.}}
  \label{fig:detection_pipeline}
\end{figure}

In the last decade, the application of detection of bleeding regions in VCE frames has witnessed significant advancements with the implementation of CNN and various other DL algorithms in this domain. The latest studies have focused on introducing DL models and improving currently released algorithms to increase the level of visual features extracted from VCE frames for detecting pathologies, including bleeding. One of the articles reviewed in this paper applied the You Only Look Once (YOLO) algorithm for detecting Multiple Bleeding Spots (MBS), while another improved upon a previously applied CNN architecture for detecting multiple pathologies to localize bleeding regions. Typically the detection pipeline includes a sequential process for identifying bleeding anomaly in frames, encompassing preprocessing, feature extraction, region proposal, feature fusion, detection, localization, post-processing, and model evaluation, ultimately outputting classified and localized instances of bleeding. Fig. \ref{fig:detection_pipeline} shows a generalized pipeline for detection method. The key differences lie in the various architectures applied.

\subsection{Classification}\label{subsec29}

Mackiewicz, Fisher, and Jamieson \cite{mackiewiz2008} proposed a bleeding detection method in VCE frames using an adaptive color histogram model and SVM. They addressed the challenge of blood color alteration due to GI fluids. It had an accuracy of 97\%. Liu and Yuan \cite{liu2008} employed SVM for obscure bleeding detection, using a dataset of 800 VCE frames, manually classified by specialists. This work highlighted the importance of color feature selection in enhancing the performance of SVM classifiers. Li and Meng \cite{li2009chroma} explored the detection of bleeding and ulcers in VCE frames using a three-layer MLP, focusing on chromaticity moments. This approach used Tchebichef and Krawtchouk discrete moments for feature extraction. Validated with VCE video data from 10 patients, the method achieved a sensitivity of 88.62\% and a specificity of 87.81\%.

In a subsequent work, Li and Meng \cite{li2009cad} introduced the concept of chrominance moment using Tchebichef polynomials and the hue/saturation/intensity (HSI) color space to detect bleeding regions in VCE frames. The method achieved about 90\% for accuracy, specificity, and sensitivity. Pan et al. \cite{pan2009bp} explored the use of backpropagation (BP) neural network for bleeding detection in VCE. They extracted color texture features in both RGB and HSI color spaces, achieving a sensitivity of 93\% and a specificity of 96\%. Further expanding the use of neural networks, Pan, Yan, Qiu, and Cui \cite{pan2011pnn} employed a Probabilistic Neural Network (PNN) to distinguish between bleeding regions from non-bleeding regions in VCE frames, achieving a sensitivity of 93.1\% and a specificity of 85.8\%.

Al-Rahayfeh and Abuzneid \cite{rahayfeh2010detection} proposed a method to classify bleeding in VCE frames. Each frame was divided into pixels, and the range ratio colour condition was applied to each pixel. If any pixels meet this condition, the frame is classified as bleeding; otherwise, it is classified as non-bleeding. Tested on 100 VCE frames, it was found that the overall accuracy improved to 98\%. Li \cite{li2010detection} introduced a novel total variation model (TV-Retinex) to address the Retinex problem, along with employing a SVM for classification of bleeding frames in VCE images, achieving a sensitivity of 96.60\% and specificity of 99.50\% using the RGB+HSI feature set. Lee and Yoon \cite{lee2011real} propose a real-time technique for bleeding detection in VCE frames. The algorithm relies on statistical analysis and bleeding spot shapes. Tested on 30 CE cases the proposed method achieved a sensitivity of 99\% and a specificity of 97\%.

Lee and Yoon \cite{lee2012improvement} in their work utilized image processing for VCE to address noise and sensitivity issues, employing variance-based filtering and feature-based weighting for blood detection. Improvement was achieved by additional preprocessing techniques in the algorithm for detecting bleeding areas. Abouelenien et al. \cite{abouelenien2013cluster} introduced a cluster-based sampling and ensemble method for bleeding detection in VCE videos. This method involved feature extraction, data rebalancing, and ensemble classification, leading to improved sensitivity and specificity. Yeh et al. \cite{yeh2014} in their paper introduced a new technique for identifying bleeding and ulcers in VCE frames, using color features. The color coherence vector (CCV) features were found to improve the efficiency of VCE image analysis, achieving accuracy (92.86\%) and sensitivity (93.64\%).

In his paper, Brzeski \cite{brzeski2014visual} introduced high-level visual features for bleeding detection in VCE frames. Evaluated on 135 VCE examinations, utilizing an SVM classifier on manually assigned feature values, the efficiency was found to be 95\% under perfect feature implementation. Nawarathna et al. \cite{nawarathna2014abnormal} proposed a novel multitexture analysis method to effectively distinguish frames with mucosal abnormalities from those without by employing a "texton histogram" of image blocks as features. Experimental results demonstrated a 92\% recall and 91.8\% specificity on VCE frames. In their work, Hassan and Haque \cite{hassan2015} employed SVM for a bleeding detection algorithm that uses textual features extracted from the magnitude spectrum of VCE frames. Their approach achieved an accuracy of 99.19\%, sensitivity of 99.41\%, and specificity of 98.95\%. 

Dmitry et al. \cite{dmitry2015development} developed an automated bleeding recognition software, following a consistent three-step process: image reading, feature identification, and result submission to an SVM algorithm. It identifies pathologies and presents them in a review-friendly format, achieving an accuracy of 94\%. The dataset used consisted of 110 frames, with 50 being healthy and 60 showing abnormalities. Priya et al. \cite{priya2015bleeding} employed SVM along with superpixel segmentation to automatically detect obscure bleeding in VCE frames. By eliminating edge-pixel influence and introducing red ratio features in the RGB color space, the proposed approach showed high improvements in sensitivity and accuracy. Usman et al. \cite{usman2016} introduced a pixel-based approach for detecting bleeding regions in VCE videos using SVM. The findings demonstrate that the method exhibits high sensitivity, specificity, accuracy, and precision in identifying various bleedings in the small colon. 

Hu et al. \cite{hu2016bleeding} used SVM for bleeding and tumor detection in VCE frames. They introduced a novel geometric image feature called local contrast-enhanced higher-order local auto-correlation (LCE-HLAC) and a pre-processing technique in the HSV color space. The system achieved 98.54\% accuracy on a dataset of VCE frames from 28 patient cases. Yuan et al. \cite{yuan2016WCE}  proposed a computer-aided system for VCE image classification into bleeding, polyp, ulcer, and normal classes. They used color scale-invariant feature transform and K-means clustering for feature extraction and visual word generation. The saliency and adaptive locality-constrained linear coding (SALLC) algorithm was applied for feature encoding. The system achieved a bleeding classification accuracy of 96.60\%. Charfi et al. \cite{charfi2018computer} introduced a novel texture extraction scheme for discriminating pathological inflammation, polyp, and bleeding regions in VCE frames. Their approach combines wavelet-based LBP for feature extraction and recognition. The technique, applied with an MLP classifier on VCE frames in RGB color space, achieved a high recognition rate. 

Bchir, Ismail, and AlZahrani \cite{bchir2019multiple} introduced a technique for automating the detection of MBS in VCE frames, utilizing unsupervised learning for clustering similar frames and supervised learning to distinguish positive (MBS-containing) from negative frames. Their method achieved a 90.92\% accuracy rate. Ding et al. \cite{ding2019} employed a CNN-based approach to evaluate small bowel CE (SB-CE) frames, validated on a large multicenter dataset. The model, designed for diagnosing abnormal SB lesions, improved sensitivity from 76.89\% to 99.90\%, increased lesion detection from 54.57\% to 70.91\%, and reduced reading time from 96.6 minutes to 5.9 minutes for detecting SB abnormalities using VCE. Diamantis et al. \cite{diamantis2019look} introduced Look-Behind Fully Convolutional Neural Network (LB-FCN), for detecting GI abnormalities like polyps, ulcers, and bleeding. LB-FCN utilizes parallel convolutional layers with different filter sizes for multi-scale feature extraction. While bleeding-only results were not isolated, LB-FCN achieved 99.72\% and 93.50\% AUC on flexible and VCE datasets, respectively.

Sathiya et al. \cite{sathiya2019} proposed an automated abnormality detection technique for VCE frames using CNN. They extracted features from image patches to enhance generality and explore optimal color space components for feature extraction and classifier design. Their approach achieves an AUC of approximately 0.8 on a dataset of 137 VCE frames. Aoki et al. \cite{aoki2020automatic} utilized a deep CNN trained on 27,847 VCE frames (6,503 blood content, 21,344 normal). Evaluation on an independent test set of 10,208 small-bowel frames (208 blood content, 10,000 normal) yielded an ROC-AUC of 0.9998 and a sensitivity, specificity, and accuracy of 96.63\%, 99.96\%, and 99.89\%, respectively. The CNN-based system outperformed SBI in individual image analysis for detecting blood content in VCE frames. Khan et al. \cite{khan2020gastrointestinal} applied TL with a ResNet101 pre-trained CNN model for classification of GI abnormalities, including ulcer, polyp, and bleeding. The extracted deep features were input into a multi-class SVM (MSVM) with a cubic kernel function, achieving a classification accuracy of 99.13\%. Though bleeding-only results were not isolated, the datasets included bleeding frames.

Khan et al. \cite{khan2020stomachnet} proposed StomachNet, an automated method for classifying GI infections, including ulcer, bleeding, polyp, and healthy classes. They utilized ResNet101 with deep TL, enhanced contrast techniques, and metaheuristic algorithms for optimal feature extraction. Fused feature vectors were input into an Extreme Learning Machine (ELM) classifier, achieving 99.46\% accuracy on a combined database. While bleeding-only results were not isolated, the datasets included bleeding frames. Pannu et al. \cite{pannu2020deep}introduced a CNN-augmented supervised learning ensemble architecture designed to diagnose bleeding symptoms in VCE frames. The proposed method achieved an accuracy of 0.95 in the public RLE data set and 0.93 in the real video data set. Shahril et al. \cite{shahril2020bleeding} evaluated a deep CNN (DCNN) for bleeding detection in VCE frames. Their pre-processing technique improved accuracy. DCNN outperformed SVM and Fuzzy logic. Specificity, sensitivity, and average accuracy were 0.8703, 0.8271, and 0.8907, respectively.

Jain et al. \cite{jain2020detection} proposed a novel approach for automatically detecting abnormalities in VCE frames using fractal features. They extract fractal dimensions (FD) using the differential box counting method and employ a RF-based ensemble classifier. The results showed AUC scores of 85\% and 99\% on the KID dataset and a private dataset, respectively. 
Kundu et al. \cite{kundu2020least} proposed a two-stage technique for classifying GI diseases using salient points of interest (POI) extracted via Least Square Saliency Transformation (LSST). A Probability Density Function (PDF) model was fitted on the POI for supervised classification, yielding high performance with a hierarchical cascaded scheme of binary SVMs. Majid et al. \cite{majid2020classification} proposed a fully automated system for recognizing gastric infections from VCE frames, including ulcer, polyp, esophagitis, and bleeding. The system involved database creation, feature extraction (discrete cosine transform, discrete wavelet transform, strong color feature, and VGG16-based CNN features), feature fusion, robust feature selection via genetic algorithm (GA), and disease recognition using an Ensemble Classifier. The system achieved 96.5\% accuracy on a database comprising four datasets, although specific bleeding-only classification results were not isolated.

Ponnusamy et al. \cite{ponnusamy2020efficient} proposed a method for classifying GI tract frames from VCE into various classes (Z-Line, Bleeding, Pylorus, Cecum, Esophagitis, Polyps, Ulcerative Colitis) using visual words. They utilized a VBOF approach with center symmetric LBP (CS-LBP), auto-color correlogram (ACC), and scale-invariant feature transform (SIFT) to capture interest points, texture, and color information. Features were clustered into visual words using K-means and classified with SVM. Their method achieved 94.80\% accuracy for bleeding detection using the Kvasir dataset. Patel et al. \cite{patel2021automated} proposed a sparse coding approach for bleeding detection in VCE frames. They utilized SIFT keypoint detector to extract patches around keypoints, reducing patch numbers significantly. SIFT and uniform LBP feature descriptors were computed around keypoints and sparse coded to obtain new feature vectors, then max-pooled for classification with SVM. The method achieved 98.18\% accuracy on a dataset of 912 frames. Mamun et al. \cite{mamun2021color} developed a computer-aided system for bleeding detection in VCE frames using color thresholding and Quantum SVM (QSVM) classification. The system achieved 95.8\% accuracy, 95\% sensitivity, 97\% specificity, and an 85\% F1 score.

Mamun et al. \cite{mamun2021} proposed a computer-automated system for bleeding detection in VCE frames using fuzzy logic technique and QSVM classifier. It incorporated fuzzy logic edge detection and statistical feature vector extraction with PCA in HSV color space. The proposed approach has been tested on 2393 VCE frames, achieving a sensitivity of 98\%, accuracy of 98.2\%, specificity of 98\%, NPV of 99\%, precision of 93\%, and F1 Score of 95.4\%. In their work, Rustam et al. \cite{rustam2021} introduce BIR (Bleedy Image Recognizer), a model combining MobileNet and a custom CNN, which was evaluated on a dataset of 1650 VCE frames, yielding accuracy, precision, recall, F1 score, and Cohen's kappa values of 0.993, 1.000, 0.994, 0.997, and 0.995, respectively. Additionally, when tested on a Google-collected VCE image dataset, BIR had an accuracy of 0.978. In their paper, Saraiva et al. \cite{saraiva2022artificial} applied CNN to detect blood and hematic residues in VCE frames. The proposed model achieved an accuracy and precision of 98. 5\% and 98. 7\%, respectively, to detect these elements in the lumen of the small intestine. The sensitivity and specificity were 98.6\% and 98.9\%, respectively.

Yogapriya et al. \cite{yogapriya2021} employed traditional image processing algorithms, data augmentation, and adjusted pretrained CNNs (VGG16, ResNet-18, GoogLeNet) to classify GI diseases in VCE frames. While specific results for bleeding-only classification were not isolated, the datasets employed in the work did contain bleeding frames. VGG16 achieved the highest performance with 96.33\% accuracy, 96.37\% recall, 96.5\% precision, 96.5\% F1-measure,  MCC (0.95), and Cohen’s kappa score (0.96). In their paper, Latha et al. \cite{latha2021deep} employed a CNN-based supervised learning ensemble to detect bleeding in VCE frames. The frames underwent preprocessing, followed by feature extraction using CNN architecture. Hyperparameter tuning involved varying parameters such as optimizers, epochs, and batch sizes. Using the Kvasir Capsule dataset, they achieved 95.7\% accuracy with 97.1\% sensitivity and 94.6\% specificity. Khan et al. \cite{khan2021blockchain} proposed a system for detecting stomach abnormalities such as ulcers and bleeding from VCE frames. They integrated blockchain-based security into a CNN model, processed frames with pre-trained deep models, fused features using a Mode value-based approach, optimized with a GA and entropy function, and employed a Softmax classifier for classification. Experimental results on a private dataset showed 96.8\% accuracy, including bleeding frames.

Nayyar et al. \cite{nayyar2021gastric} utilized hybrid contrast enhancement and fine-tuned a pre-trained AlexNet model with TL. Deep features were fused using a vector length-based approach and optimized with a genetic algorithm. Evaluation on 24,000 VCE frames achieved 99\% average accuracy using a cubic SVM (CSVM) classifier. The validation included datasets from various sources such as Kvasir and CVC. Modi et al. \cite{modi2021digestive} proposed a system achieving 97.82\% accuracy on the Kvasir-capsule dataset for training and predicting 13 classes of digestive tract anomalies. They preprocessed the dataset for uniform image sizes and used the ImageDataGenerator library for label matching. Their CNN Sequential model effectively detected abnormalities, with the RMSprop optimizer improving accuracy over Adadelta, despite dataset imbalance. Kim et al. \cite{kim2021efficacy} employed the Inception-ResNet-V2 model for binary classification to selectively identify inflamed mucosa, atypical vascularity, or bleeding in VCE frames, achieving 98\% diagnostic accuracy and an AUC of 0.99 during internal testing. However, external testing showed a 10\% accuracy decrease and a 0.07 AUC drop. Although specific results for bleeding-only classification were not isolated, the datasets employed in the work contained bleeding frames. This was the first AI model developed and tested using the Mirocam capsule image-set.

\begin{figure}[ht]
  \centering
  \includegraphics[width=\columnwidth]{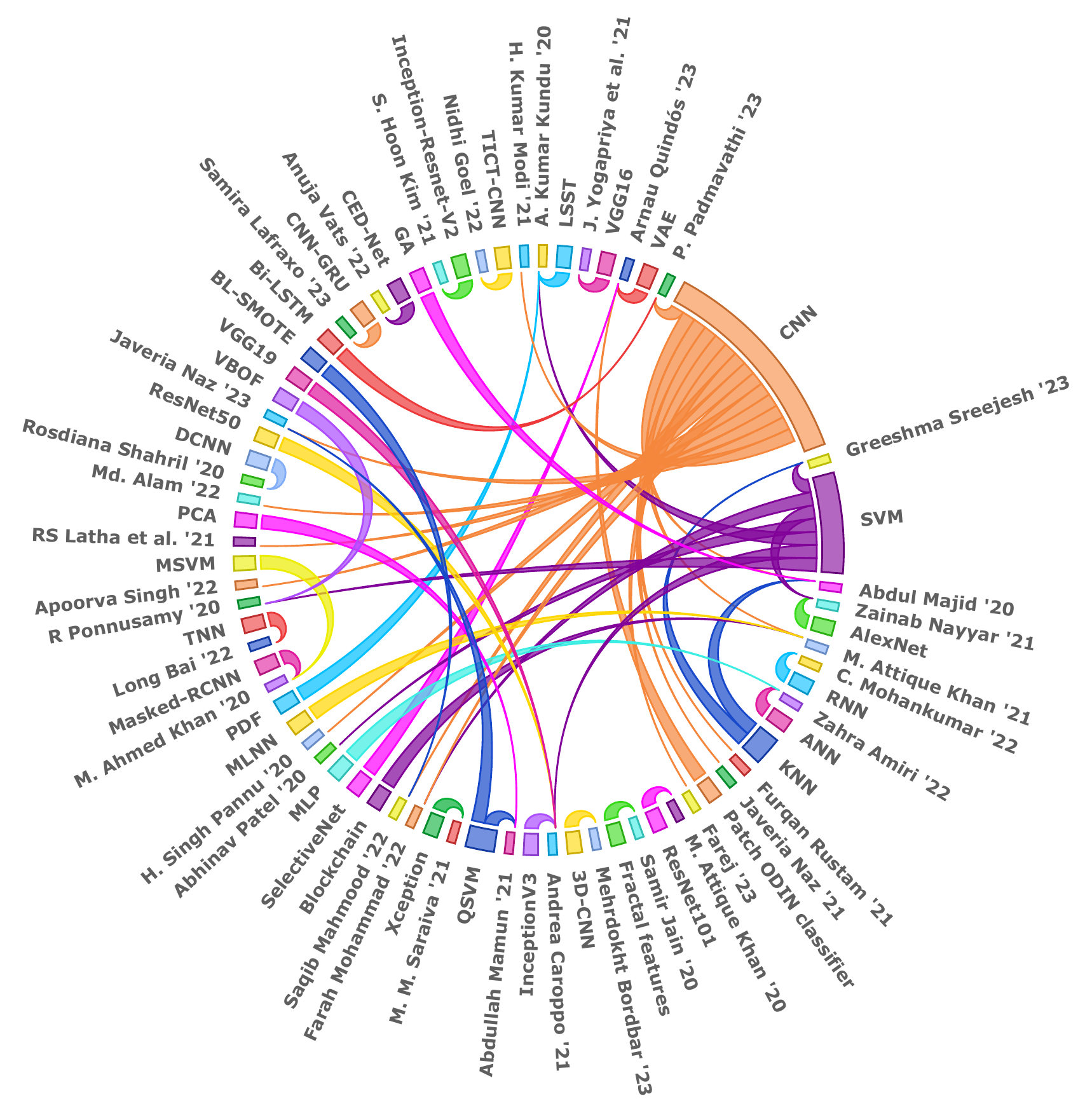}
  \caption{\RaggedRight{Authors and Methods: Chord Chart for Bleeding Frame Classification in CE Images.}}
  \label{fig:Classification_chordchart}
\end{figure}

Khan et al. \cite{khan2021multiclass} proposed a method for multiclass classification of stomach diseases using VCE frames. It involved preprocessing, deep TL-based feature extraction with pretrained Inception V3, and feature optimization using PSO and CSA. Classification was done with a Multi-Layered Perceptron Neural Network (MLNN). Evaluation on two datasets (CUI WahStomach Diseases and Combined dataset) showed an average accuracy of 99.5\%. While specific results for bleeding-only classification were not isolated, the datasets employed in the work did contain bleeding frames. Naz et al. \cite{naz2021recognizing} introduced a hybrid method combining LBP and Segmentation-based Fractal Texture Analysis (SFTA) texture features with pretrained CNN features (VGG16 and InceptionV3). The enhanced image dataset undergoes power-law transformation to improve contrast. The system achieved classification accuracies of 99.25\%, 99.75\%, and 100\% for the KVASIR, Nerthus, and Ulcer-Bleeding datasets, respectively. Caroppo et al. \cite{caroppo2021deep} utilized three pre-trained deep CNNs (VGG19, InceptionV3, ResNet50) for feature extraction and fusion was done using the minimum Redundancy Maximum Relevance (mRMR) method. Bleeding/non-bleeding VCE frame classification was done with SVM. Evaluation on KID and MICCAI 2017 datasets yielded average accuracies of 97.65\% and 95.70\% for bleeding region detection.

In their work, Alam et al. \cite{alam2022} proposed RAt-CapsNet, a CNN architecture designed to classify abnormalities in VCE video data. By leveraging regional information and attention mechanisms, the proposed model achieves a bleeding accuracy of 100\% on the Kvasir public dataset. Amiri et al. \cite{amiri2022} used an MLP classifier to analyze VCE frames for anomalies like bleeding and angiodysplasia lesions. Their method identifies regions of interest and extracts features using a combination of color histogram analysis and statistical features. It achieved precision and recall rates of approximately 96.5\% and 95.9\%, respectively. While specific results for bleeding-only classification were not isolated, the datasets used in the work did contain bleeding frames. Vats et al. \cite{vats2022} employed a multichannel encoder-decoder network to learn pathology patterns in VCE frames from discriminative cues like the presence of bleeding indicated by red. Despite using fewer than 2500 labels for training, the approach accurately classified various pathologies in two different VCE datasets, namely Kvasir-capsule and Computer-Assisted Diagnosis for CAPsule Endoscopy database (CAD-CAP), achieving specificity ($>$97\%) for all classes.

Mohankumar et al. \cite{mohankumar2022} employed RNN DL to identify GI abnormalities in VCE frames, achieving 93.27\% accuracy. The work included ulcers and bleeding detection, utilizing MLPs for local categorization and experimenting with color spaces for feature extraction. While specific bleeding-only classification results were not isolated, the datasets used contained bleeding frames. Bai et al. \cite{bai2022} employed a transformer neural network (TNN) with spatial pooling to train on two publicly available CE disease datasets for bleeding detection. They achieved 79.15\% accuracy on the Kvasir-Capsule multi-classification task and 98.63\% bleeding accuracy on the RLE binary classification task. Singh et al. \cite{singh2022explainable} utilized a CNN model to classify bleeding frames in a VCE video dataset, extracted using VLC software. The approach included pre-processing for standardization and denoising. The method achieved a test accuracy of 91.92\%, sensitivity of 68.42\%, specificity of 97.48\%, precision of 86.67\%, and F1 score of 0.7647 on private dataset comprising 2,621 frames (505 bleeding, 2,116 normal). Additionally, they proposed an explainable tool for medical experts to detect gastroenterological bleeding more efficiently and reliably.

\begin{table*}[htbp]
\centering
\caption{Papers Implementing Bleeding Frame Classification.}
\label{table: classification_table1}
\scriptsize
\begin{adjustbox}{width=\textwidth,totalheight=\textheight,keepaspectratio}
\renewcommand{\arraystretch}{1.2}
\setlength{\tabcolsep}{4pt}
\begin{tabular}{|>{\RaggedRight\arraybackslash}p{2.7cm}|>{\RaggedRight\arraybackslash}p{1.5cm}|>{\RaggedRight\arraybackslash}p{2.1cm}|>{\RaggedRight\arraybackslash}p{1.7cm}|>{\RaggedRight\arraybackslash}p{1.4cm}|>{\RaggedRight\arraybackslash}p{2.8cm}|}
\hline
\textbf{Reference} & \textbf{Year} & \textbf{Technique} & \textbf{Dataset} & \textbf{Accuracy} & \textbf{Other Evaluation Metrics} \\
\hline
Mackiewicz et al.~\cite{mackiewiz2008} & 2008 & SVM & Frames from 10 VCE videos & 97\% & - \\
\hline
Liu et al.~\cite{liu2008} & 2008 & SVM & 800 VCE frames & - &Sensitivity - 99\%, Specificity - 99\% \\
\hline
Li et al.~\cite{li2009chroma} & 2009 & Three-Layer MLP & 100 VCE frames & - & Sensitivity - 88.62\%, Specificity - 87.81\% \\
\hline
Li et al.~\cite{li2009cad} & 2009 & Three-Layer MLP & - & 90\% & Sensitivity - 90\%, Specificity - 90\% \\
\hline
Pan et al.~\cite{pan2009bp} & 2009 & RGB/HSI + BP & Train - 20 VCE frames, Test - 150 VCE video & - & Sensitivity - 93\%, Specificity - 96\% \\
\hline
Pan et al.~\cite{pan2011pnn} & 2010 & PNN & Image level test - 150 VCE video & - & Sensitivity - 93.1\%, Specificity - 85.8\% on image level \\
\hline
Al-Rahayfeh et al.~\cite{rahayfeh2010detection} & 2010 & Range ratio color & Test - 100 VCE frames & 98\% & - \\
\hline
Li~\cite{li2010detection} & 2010 & TV-Retinex, SVM & 1150 frames & - & Sensitivity - 96.6\%, Specificity - 99.5\% \\
\hline
Lee et al.~\cite{lee2011real} & 2011 & Statistical analysis, bleeding spot shapes & 30 VCE cases/3000 frames & - & Sensitivity - 99\%, Specificity - 97\% \\
\hline
Lee et al.~\cite{lee2012improvement} & 2012 & Image processing & 30 VCE frames & - & - \\
\hline
Abouelenien et al.~\cite{abouelenien2013cluster} & 2013 & SVM & Frames from 8 annotated VCE videos & - & - \\
\hline
Yeh et al.~\cite{yeh2014} & 2014 & Using color features & Capsuleend\-oscopy.org,\- RAPID Atlas & 92.86\% & Sensitivity - 93.64\% \\
\hline
Brzeski~\cite{brzeski2014visual} & 2014 & SVM & 125 VCE frames & 95\% & - \\
\hline
Nawarathna et al.~\cite{nawarathna2014abnormal} & 2014 & Texton histogram & - & - & Recall - 92\%, Specificity - 91.8\% \\
\hline
Hassan et al.~\cite{hassan2015} & 2015 & SVM & 1200 VCE frames & 99.19\% & Sensitivity - 99.41\%, Specificity - 98.95\% \\
\hline
Dmitry et al.~\cite{dmitry2015development} & 2015 & SVM & 110 VCE frames & 94\% & - \\
\hline
Priya et al.~\cite{priya2015bleeding} & 2015 & SVM & - & - & - \\
\hline
Usman et al.~\cite{usman2016} & 2016 & SVM & 8500 VCE frames & - & - \\
\hline
Hu et al.~\cite{hu2016bleeding} & 2016 & SVM & 28 cases & 98.54\% & - \\
\hline
Yuan et al.~\cite{yuan2016WCE} & 2016 & SALLC algorithm & 1650 VCE frames & 96.60\% & - \\
\hline
Charfi et al.~\cite{charfi2018computer} & 2018 & Wavelet-based LBP, MLP & - & - & - \\
\hline
Bchir et al.~\cite{bchir2019multiple} & 2019 & SVM, KNN & 1275 frames & 90.9\% & - \\
\hline
Ding et al.~\cite{ding2019} & 2019 & CNN & 113,426,569 frames from 6970 patients & - & Sensitivity - 99.88\% \\
\hline

\end{tabular}
\end{adjustbox}
\end{table*}

\begin{table*}[htbp]
\centering
\label{table: classification_table2}
\scriptsize
\begin{adjustbox}{width=\textwidth,totalheight=\textheight,keepaspectratio}
\renewcommand{\arraystretch}{1.2}
\setlength{\tabcolsep}{4pt}
\begin{tabular}{|>{\RaggedRight\arraybackslash}p{2.7cm}|>{\RaggedRight\arraybackslash}p{1.5cm}|>{\RaggedRight\arraybackslash}p{2.1cm}|>{\RaggedRight\arraybackslash}p{1.7cm}|>{\RaggedRight\arraybackslash}p{1.4cm}|>{\RaggedRight\arraybackslash}p{2.8cm}|}
\hline
\textbf{Reference} & \textbf{Year} & \textbf{Technique} & \textbf{Dataset} & \textbf{Accuracy} & \textbf{Other Evaluation Metrics} \\
\hline
Diamantis et al.~\cite{diamantis2019look} & 2019 & LB-FCN & MICCAI 2015, KID & - & 99.72\% AUC \\
\hline
Sathiya et al.~\cite{sathiya2019} & 2019 & CNN & 137 VCE frames & - & AUC - 0.8 \\
\hline
Aoki et al.~\cite{aoki2020automatic} & 2019 & CNN & Train - 27,847 VCE frames, Test: 10,208 VCE frames & 99.89\% & ROC-AUC - 0.9998, Sensitivity - 96.63\%, Specificity - 99.96\% \\
\hline
Khan et al.~\cite{khan2020gastrointestinal} & 2020 & CNN, MSVM & Frames from 30 VCE videos & 99.13\% & - \\
\hline
Khan et al.~\cite{khan2020stomachnet} & 2020 & CNN & - & 99.46\% & - \\
\hline
Pannu et al.~\cite{pannu2020deep} & 2020 & CNN & RLE Dataset, Private Dataset & 95\% on public dataset, 93\% on real video dataset & - \\
\hline
Shahril et al.~\cite{shahril2020bleeding} & 2020 & DCNN & - & 89.07\% & Specificity - 87.03\%, Sensitivity - 82.71\% \\
\hline
Jain et al.~\cite{jain2020detection} & 2020 & Fractal features & KID, Private dataset & - & AUC scores of 85\%, 99\% on KID, Private dataset \\
\hline
Kundu et al.~\cite{kundu2020least} & 2020 & LSST, PDF, SVM & 2588 VCE frames & - & - \\
\hline
Majid et al.~\cite{majid2020classification} & 2020 & CNN, GA, KNN & \tiny{Kvasir-Capsule, CVC-ClinicDB, Private, and \-ETIS Larib\-PolypDB} & 96.5\% & - \\
\hline
Ponnusamy et al.~\cite{ponnusamy2020efficient} & 2020 & VBOF, SVM & Kvasir-Capsule & 94.80\% & - \\
\hline
Patel et al.~\cite{patel2021automated} & 2020 & SVM & 912 VCE frames & 98.18\% & - \\
\hline
Mamun et al.~\cite{mamun2021color} & 2021 & QSVM & 2393 annotated VCE frames & 95.8\% & Sensitivity - 95\%, Specificity - 97\%, Precision - 80\%, NPV - 99\%, F1 score - 85\% \\
\hline
Mamun et al.~\cite{mamun2021} & 2021 & QSVM, PCA & 2393 annotated VCE frames & 98.2\% & Sensitivity - 98\%, Specificity - 98\%, NPV - 99\%, Precision - 93\%, F1 Score - 95.4\% \\
\hline
Rustam et al.~\cite{rustam2021} & 2021 & CNN & Frames from 33 patients' VCE videos & 99.3\% & Precision - 100\%, Recall - 99.4\%, F1 score - 99.7\%, Cohen's kappa - 99.5\% \\
\hline
Saraiva et al.~\cite{saraiva2022artificial} & 2021 & CNN & 22,095 VCE frames & 98.5\% & Precision - 98.7\%, Sensitivity - 98.6\%, Specificity - 98.9\% \\
\hline
Yogapriya et al.~\cite{yogapriya2021} & 2021 & CNN & Kvasir-Capsule & 96.33\% & Recall - 96.37\%, Precision - 96.5\%, F1-measure - 96.5\%, MCC - 95\%, Cohen’s kappa score - 96\% \\
\hline

\end{tabular}
\end{adjustbox}
\end{table*}
\raggedbottom

\begin{table*}[htbp]
\centering
\label{table: classification_table3}
\scriptsize
\begin{adjustbox}{width=\textwidth,totalheight=\textheight,keepaspectratio}
\renewcommand{\arraystretch}{1.2}
\setlength{\tabcolsep}{4pt}
\begin{tabular}{|>{\RaggedRight\arraybackslash}p{2.7cm}|>{\RaggedRight\arraybackslash}p{1.5cm}|>{\RaggedRight\arraybackslash}p{2.1cm}|>{\RaggedRight\arraybackslash}p{1.7cm}|>{\RaggedRight\arraybackslash}p{1.4cm}|>{\RaggedRight\arraybackslash}p{2.8cm}|}
\hline
\textbf{Reference} & \textbf{Year} & \textbf{Technique} & \textbf{Dataset} & \textbf{Accuracy} & \textbf{Other Evaluation Metrics} \\
\hline
Latha et al.~\cite{latha2021deep} & 2021 & CNN & Kvasir-Capsule & 95.7\% & Sensitivity - 97.1\%, Specificity - 94.6\% \\
\hline
Khan et al.~\cite{khan2021blockchain} & 2021 & CNN, Blockchain & Private dataset & 96.8\% & - \\
\hline
Nayyar et al.~\cite{nayyar2021gastric} & 2021 & CNN, SVM & 24,000 VCE frames from Kvasir-Capsule, CVC, and Private dataset & 99\% & - \\
\hline
Modi et al.~\cite{modi2021digestive} & 2021 & CNN & Kvasir-Capsule & 97.83\% & - \\
\hline
Kim et al.~\cite{kim2021efficacy} & 2021 & CNN & Mirocam capsule image set & 98\% & AUC - 0.99 \\
\hline
Khan et al.~\cite{khan2021multiclass} & 2021 & CNN, MLNN & CUI WahStomach Diseases, Combined dataset & 99.5\% & - \\
\hline
Naz et al.~\cite{naz2021recognizing} & 2021 & CNN & Kvasir-Capsule, Nerthus, and Ulcer-Bleeding dataset & 99.3\%, 99.9\%, 100\% & - \\

\hline
Caroppo et al.~\cite{caroppo2021deep} & 2021 & CNN, SVM & KID (2352 frames) and MICCAI 2017 (3895 frames) & 97.65\%, 95.70\% & - \\
\hline
Alam et al.~\cite{alam2022} & 2022 & CNN & Kvasir-Capsule & 100\% & - \\
\hline
Amiri et al.~\cite{amiri2022} & 2022 & MLP, ANN & GIANA 2017 dataset, 600 frames & - & Precision - 96.5\%, Recall - 95.9\% \\
\hline
Vats et al. \cite{vats2022}& 2022 & Cascaded Encoder-Decoder network & Kvasir-Capsule, CAD-CAP & - & Specificity - 97\%\\
\hline
Mohankumar et al. \cite{mohankumar2022}& 2022 & RNN & Private Dataset & 93.27\%& -\\
\hline
Bai et al. \cite{bai2022}& 2022 & TNN & Kvasir-Capsule, RLE & 79.15\% in Multi and 98.63\% in binary classification & - \\
\hline
Singh et al. \cite{singh2022explainable}& 2022 & CNN &  2,621 VCE frames & 91.92\% & Sensitivity - 68.42\%, Specificity - 97.48\%, Precision - 86.67\%, F1 score - 76.47\%\\
\hline
Goel et al. \cite{goel2022investigating}& 2022 & TICT-CNN & AIIMS Delhi dataset, KID &-& -.\\
\hline
Mahmood et al. \cite{mahmood2022robust}& 2022 & CNN, BL-SMOTE & Kvasir-Capsule& 98.9\% & AUC - 99.8\%, F1-score - 98.9\%, Precision - 98.9\%, Recall - 98.8\% , Loss - 0.0474\\
\hline
\end{tabular}
\end{adjustbox}
\end{table*}

\begin{table*}[htbp]
\centering
\label{table: classification_table5}
\scriptsize
\begin{adjustbox}{width=\textwidth,totalheight=\textheight,keepaspectratio}
\renewcommand{\arraystretch}{1.2}
\setlength{\tabcolsep}{4pt}
\begin{tabular}{|>{\RaggedRight\arraybackslash}p{2.7cm}|>{\RaggedRight\arraybackslash}p{1.5cm}|>{\RaggedRight\arraybackslash}p{2.1cm}|>{\RaggedRight\arraybackslash}p{1.7cm}|>{\RaggedRight\arraybackslash}p{1.4cm}|>{\RaggedRight\arraybackslash}p{2.8cm}|}
\hline
\textbf{Reference} & \textbf{Year} & \textbf{Technique} & \textbf{Dataset} & \textbf{Accuracy} & \textbf{Other Evaluation Metrics} \\
\hline
Mohammad et al. \cite{mohammad2022deep}& 2022 & CNN, SVM & - & 99.8\%& -\\
\hline
Lafraxo et al. \cite{lafraxo2023computer}& 2023 & CNN-GRU & MICCAI 2017&- & Precision - 99. 39\%\\
\hline
Farej et al. \cite{farhan2023accuracy}& 2023 & CNN & Kvasir-Capsule& 99.44\%  & Precision - 99.47\%, Recall - 99.412\%, F1 score - 99.44\%\\
\hline
Bordbar et al. \cite{bordbar2023wireless}& 2023 & 3D-CNN & Frames from 29 VCE videos & 99.20\%& Sensitivity - 98.92\%, Specificity - 99.50\%\\
\hline
Naz et al. \cite{naz2023comparative}& 2023 & CNN, QSVM & Hybrid dataset, Kvasir-V1 dataset & 100\%, 99.24\% & -\\
\hline
Sreejesh \cite{sreejesh2023bleeding} & 2023 & SVM, KNN & - & 95.75\% & Sensitivity - 92\%, Specificity - 96.5\%, AUC - 0.9771 \\
\hline
Quindós et al. \cite{quindos2023}& 2023 & Patch ODIN classifier, SelectiveNet, VAE & Kvasir-Capsule & - & AUROC $>$ 0.6\\
\hline
Padmavathi et al. \cite{padmavathi2023wireless}& 2023 & CNN, Bi-LSTM & - & 99.6\% & - \\
\hline
Kaur et al. \cite{kaur2023wireless}& 2023 & CNN, RF & WCE Curated Colon Disease Dataset & 98.75\% & F-Score - 98\%\\
\hline
\end{tabular}
\end{adjustbox}
\end{table*}

Goel et al. \cite{goel2022investigating} employed Transforming Input Color Space in Tandem with CNN (TICT-CNN), a CNN-based framework, for binary VCE frame classification. It conducted on-the-fly data augmentation and color space conversion before training. They analyzed color space performance using objective parameters and feature maps, using AIIMS Delhi and KID datasets. Results favored HSV color space, reducing parameters by 2 to 6 times and achieving a diagnosis time of 0.02 s/frame. Mahmood et al. \cite{mahmood2022robust} proposed GIDD-Net, a custom CNN for efficient GI tract disorder detection from VCE frames. They balanced datasets using BL-SMOTE and visualized class activation with Grad-CAM. Evaluation on Kvasir-Capsule yielded 98.9\% accuracy, 99.8\% AUC, 98.9\% F1-score, 98.9\% precision, 98.8\% recall, and 0.0474 loss. While bleeding-only results weren't isolated, the datasets included bleeding frames. Mohammad and Al-Razgan \cite{mohammad2022deep} proposed a method for stomach disease recognition from VCE frames, classifying five classes: polyp, ulcerative colitis, esophagitis, bleeding, and healthy. The approach involved data augmentation, deep TL with Inception v3 and DenseNet-201, feature fusion, optimization with a modified dragonfly optimization method, and final classification using various ML algorithms. CSVM achieved the highest accuracy of 99.8\%. While bleeding-only results weren't isolated, the datasets included bleeding frames.

\small
Lafraxo, Ansari, and Koutti \cite{lafraxo2023computer} presented a computer-assisted technique to identify bleeding from VCE frames, involving preprocessing, optimized CNN feature extraction, and classification using Gated Recurrent Unit (GRU). The method achieved 99.39\% precision with the MICCAI 2017 dataset. Farej et al. \cite{farhan2023accuracy} developed and assessed a CNN model for VCE frame analysis, distinguishing small intestine bleeding from normal cases. Utilizing the Kvasir-Capsule database with 892 frames, the model achieved macro-average accuracy, precision, recall, and F1 scores of 0.99441, 0.99474, 0.99412, and 0.99440 respectively. Bordbar et al. \cite{bordbar2023wireless} introduced an automatic multiclass classification system based on a 3D-CNN for VCE diagnosis. The proposed 3D model outperformed pre-trained networks using 29 VCE videos (14,691 frames), achieving higher sensitivity (98.92 vs. 98.05), specificity (99.50 vs. 86.94), and accuracy (99.20 vs. 92.60). While specific results for bleeding-only classification were not isolated, the datasets contained bleeding frames.
\normalsize

Naz et al. \cite{naz2023comparative} compare optimization algorithms for classifying GI abnormalities using ensemble Xception Network (XcepNet23) and ResNet18 features. Their work includes dataset augmentation, image preprocessing, feature engineering, and optimization with metaheuristic algorithms (BDA, MFO, PSO). The QSVM achieves 100\% accuracy on the Hybrid dataset (five classes) and 99.24\% on Kvasir-V1 (eight classes). Sreejesh \cite{sreejesh2023bleeding} proposed a novel color feature extraction method to distinguish bleeding frames from normal ones and localize bleeding regions. K-means clustering characterized color histograms, and SVM and KNN performed classification. Experimental results using the YCbCr color space, 80 clusters, and SVM achieved 95.75\% accuracy and 0.9771 AUC. Quindos et al. \cite{quindos2023} introduced a patch-based self-supervised approach for analyzing VCE frames. Using a triplet network, they improved out-of-distribution (OOD) detection. Tested on the Kvasir-capsule dataset, the approach achieved AUROC \> 0.6 for detecting anomalies like lymphangiectasia, foreign bodies, and blood.

\small
Padmavathi et al. \cite{padmavathi2023wireless} introduced a method for VCE bleeding frame detection and classification using a deep neural network. Their approach included a MobileNetV2-based BIR for initial processing, followed by further analysis with a CNN. Additionally, a Bidirectional Long Short-Term Memory (Bi-LSTM)  with attention mechanism was employed to boost performance, achieving an accuracy of 0.996 with data augmentation. Kaur and Kumar \cite{kaur2023wireless} applied TL and an RF classifier with MobileNetV2 for feature extraction. They addressed challenges like small datasets and overfitting by optimizing hyperparameters through Bayesian optimization. Achieving a disease prediction accuracy of 98.75\% and reducing video viewing time, their method obtained an average F-Score of 0.98 using the WCE Curated Colon Disease Dataset. A chord chart describing the techniques used by the authors of these papers has been provided in Fig. \ref{fig:Classification_chordchart}. It explains that CNN is the most commonly used DL algorithm for classifying bleeding frames in VCE images, and SVM is a frequently implemented ML technique in this domain. Several modifications to SVM and novel CNN algorithms have also been introduced for more efficient and accurate results. The Tables \ref{table: classification_table1}, \ref{table: classification_table2}, \ref{table: classification_table3}, \ref{table: classification_table5} shows the important details of the papers reviewed in this sub-section. 
\normalsize


\subsection{Segmentation}\label{subsec30}

Pan Xu et al. \cite{pan2011novelcolor} proposed a novel method to calculate color similarity vector coefficients using the grey intensity similarity coefficient and chroma similarity coefficient. Upon utilizing this method in the segmentation of bleeding frames with 97\% sensitivity and specificity of 90\%. Yong-Gyu Lee and Gilwon Yoon \cite{lee2011bleeding} studied the weights between the contrast and intensity of the bleeding and non-bleeding VCE frames. To give more contrast to bleeding regions, the red-rated pixels were assigned more weight, and the blood pixels were compared with the neighboring ones. While testing this method, the results showed sensitivity and specificity of 87\% and 90\%, respectively. Szcyzypinsk et al. \cite{szczpinski2012} explored the differentiating effects of color and texture-based characteristics of VCE frames and proposed a new algorithm for feature selection, Vector Support Convex Hull (VSCH). VSCH gave specificity and sensitivity ratios of 98\% each in the detection of bleeding in VCE frames.

Figueiredo et al. \cite{figueiredo2013computer} introduced four bleeding detector functions (BDFs) based on the a-channel component of the CIE lab color space and enhancement techniques like adaptive anisotropic diffusion on VCE frames. These detectors rely on the eigenvalues of the Hessian and the Laplacian of the modified frames and achieved sensitivity and accuracy rates: 89.3\% and 90.4\%, for the bleeding detector 1, 92.9\% and 91.1\%, for detector 2, 92.9\% and 92.7\%, for detector 3 and 92.9\% and 91.1\%, for the bleeding detector 4, respectively. In their work, Sainju et al. \cite{sainju2014} presented an algorithm to annotate bleeding regions from successive video frames, which reduced the number of features per region for pattern recognition purposes, focusing on the fact that consecutive frames of a VCE have fewer differences. This method achieved accuracy, specificity, and sensitivity of 98.1\%, 99.3\%, and 100\%, respectively. Ramaraj et al. \cite{Ramaraj2014segmentation} introduced histogram variance-controlled bleeding detectors for the detection of bleeding regions. They contain three modules: determining histogram statistics of CIE-Lab components, defining BDFs based on these statistics, and applying K-means clustering for the segmentation of bleeding regions. The paper showed notable enhancement in detecting and segmenting bleeding regions from VCE frames, supported by ROC curves for binary classifier.

\begin{figure}[ht]
  \centering
  \includegraphics[width=\columnwidth]{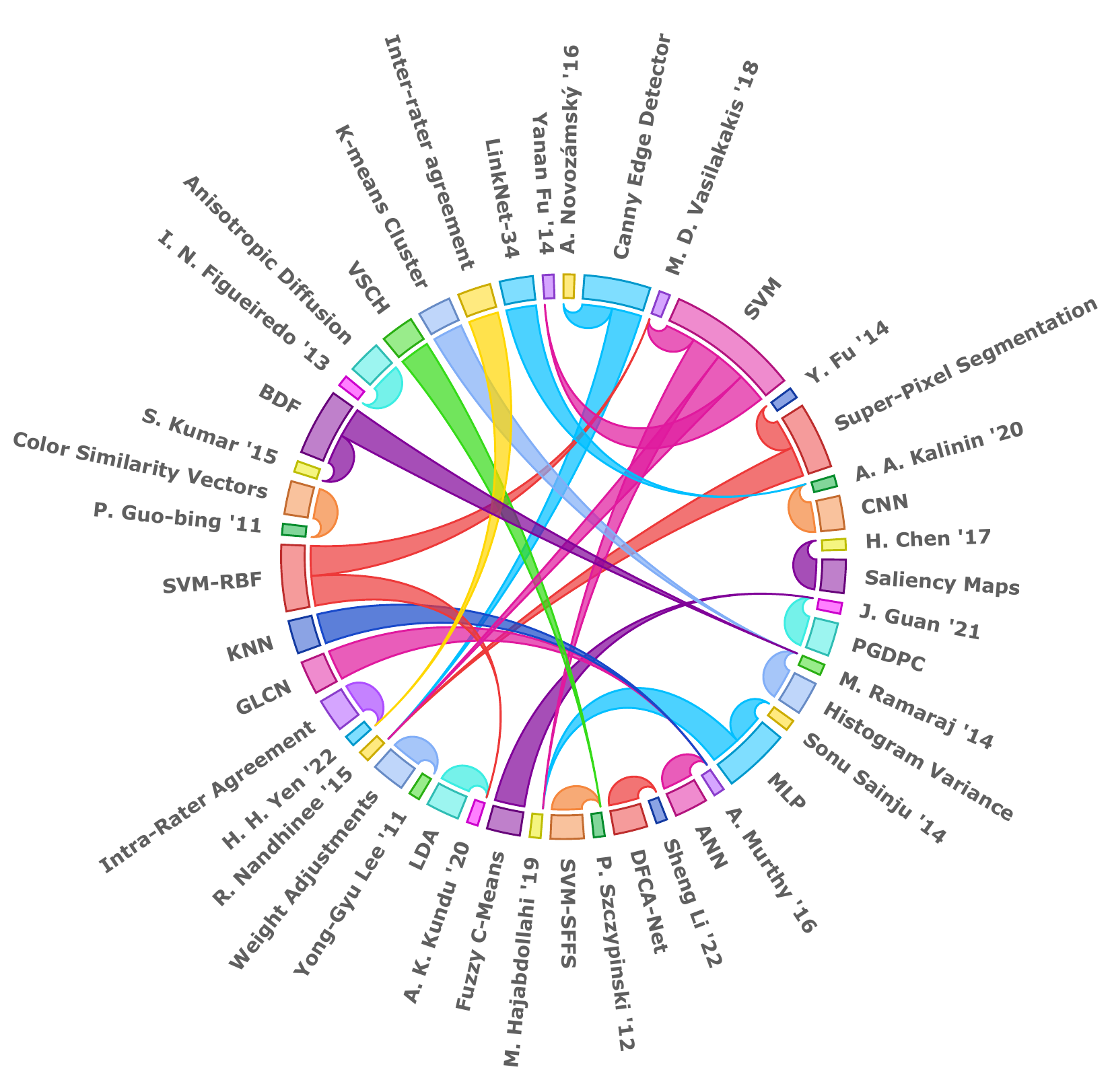}
  \caption{\RaggedRight{Chord Chart of Authors and Techniques in Bleeding Frame Segmentation for CE Images.}}
  \label{fig: Segmenatation_chordchart}
\end{figure}

Yanan Fu et al. \cite{fu2014} introduced a new method of bleeding detection in VCE frames by grouping pixels based on color and location using super-pixel segmentation. Features were extracted using the red ratio in RGB color space, minimizing the influence of edge pixels. These features were fed into SVM for classification, which resulted in a specificity of 99\%, sensitivity of 94\%, and 95\% accuracy for bleeding detection. Kumar et al. \cite{Kumar2014} introduced an algorithm that reduced the computational time of segmentation and detection of bleeding by 50 times to 72 frames per second by utilization of GPU-based parallel programming. They parallelized segmentation and blood detector function processes and presented a blood discriminator based on Hessian eigenvalues and color channel analysis. This algorithm showed an accuracy of 89.56\%. Nandhinee et al. \cite{nandhineedetection} proposed an algorithm to detect bleeding regions from VCE videos using the super-pixel segmentation process, emphasizing color and spatial similarity. The pixels are grouped based on color and location for classification using SVM classifier. The paper lacks quantitative figures to support the claim to have achieved high rates of accuracy, sensitivity and specificity. 

Novozámský et al. \cite{novozamsky2016automatic} proposed two methods for automatic bleeding detection in VCE videos: one based on color information and the other one on characteristics of blood spot, shape and size. Evaluation on private datasets revealed the optimal results when there were five suspected pixels in one frame, yielding true positive values of 96.76\% for the first method, 86.73\% for the second method, and 85.84\% when both methods were used to classify bleeding frames. A. Murthy et al. \cite{murthi2016automatic} presented a feature extraction method for detection of blood in VCE frames which gave an accuracy of 97.25\%. This method included detection of bleeding by utilizing a word-based histogram, implementation of Gray Level Co-occurrence Matrix (GLCM) using ANN and KNN for classification of bleeding frames. Further, implementation of a two-stage saliency map extraction method to observe bleeding regions under different color components and identify them by the red color. Chen et al. \cite{chen2017saliency} proposed a system for the localization of bleeding regions in VCE frames by computing three saliency maps: an edge saliency map based on phase congruency from Log-Gaber filter bands, another based on intensity histogram, and the last one based on the red proportions of the image. These three maps are fused together and threshold is generated using which the bleeding regions are localized.

Vasilakakis et al. \cite{vasilakakis2018} presented a feature extraction method by applying unsupervised color-based saliency detection, DIstaNces On Selective Aggregation of chRomatic image Components (DINOSARC). It includes saliency at point and region-level as well as color information on local and global levels, which resulted in fewer and more significant saliency points for the detection of bleeding regions in VCE frames. The AUC and accuracy achieved with DINOSARC were 81.3\% and 80.9\%, respectively, with local image descriptors and 81.5\% and 81.8\%, respectively, with global image descriptors. Hajabdollahi et al. \cite{hajabdollahi2019} proposed simplification methods in CNN and MLP to reduce the computational complexity of neural networks used in the automatic segmentation of bleeding regions in VCE frames. Upon evaluating their individual performance, it was found that the simplified CNN needed almost half of the parameters while having an AUC of ROC exceeding 97\%, and it was better than simplified MLP. However, modified MLP required even smaller computation operations to segment bleeding regions. A.A. Kalinin et al. \cite{kalinin2020} investigated the use of deep CNNs for the segmentation of angiodysplasia lesions in VCE frames. They integrated ImageNet pre-trained encoders into the U-net architecture for improved performance. It was found that models with ImageNet encoders, i.e., pre-trained VGG-11, VGG-16, and ResNet-34 networks, outperform U-Net trained from scratch, achieving an IOU of 75.35\%, DICE score of 84.98\%. The paper also presented the model's performance on multi-class instrument segmentation, but we have focused on bleeding only.  

\begin{table*}[htbp]
\centering
\caption{Papers Implementing Bleeding Frame Segmentation.}
\label{table: segmentation_table1}
\scriptsize
\begin{adjustbox}{width=\textwidth,totalheight=\textheight,keepaspectratio}
\renewcommand{\arraystretch}{1.2}
\setlength{\tabcolsep}{4pt}
\begin{tabular}{|>{\RaggedRight\arraybackslash}p{2.7cm}|>{\RaggedRight\arraybackslash}p{1.5cm}|>{\RaggedRight\arraybackslash}p{2.1cm}|>{\RaggedRight\arraybackslash}p{1.7cm}|>{\RaggedRight\arraybackslash}p{1.4cm}|>{\RaggedRight\arraybackslash}p{2.8cm}|}
\hline
\textbf{Reference} & \textbf{Year} & \textbf{Technique} & \textbf{Dataset} & \textbf{Accuracy} & \textbf{Other Evaluation Metrics} \\
\hline
Pan et al. \cite{pan2011novelcolor} & 2011 & Color similarity vectors & 20 VCE bleeding frames & - & Sensitivity - 90\%, Specificity - 97\% \\
\hline
Yong-Gyu Lee et al. \cite{lee2011bleeding} & 2011 & Color-based weights adjustment & 42 VCE frames & - & Sensitivity - 87\%, Specificity - 90\% \\
\hline
Szczypinsk et al. \cite{szczpinski2012} & 2012 & VSCH, SVM-SFFS & 50 VCE videos & - & Recall - 0.875, Precision - 0.732, Jaccard Index - 0.48 \\
\hline
Figueirido et al. \cite{figueiredo2013computer} & 2013 & Anisotropic diffusion & 700, 1200, and 2087 VCE frames from 3 datasets & 92.7\% & Sensitivity - 92.9\% \\
\hline
Sainju et al. \cite{sainju2014} & 2014 & Two-layer feed-forward MLP & 100 VCE frames & 98.1\% & Sensitivity - 100\%, Specificity - 99.3\% \\
\hline
Ramaraj et al. \cite{Ramaraj2014segmentation} & 2014 & Histogram variance, BDF, K-means clustering & - & - & - \\
\hline
Yanan Fu et al. \cite{fu2014} & 2014 & Superpixel segmentation, SVM & 5000 VCE frames & 95\% & Specificity - 94\%, Sensitivity - 99\% \\
\hline
Kumar et al. \cite{Kumar2014} & 2015 & Parallelizing segmentation and BDF & 690 VCE frames & 89.56\% & - \\
\hline
Nandhinee et al. \cite{nandhineedetection} & 2015 & Canny Edge Detector, Superpixel segmentation, SVM & - & - & - \\
\hline
Novozámský et al. \cite{novozamsky2016automatic} & 2016 & Canny Edge Detector, Morphological Erosion & 12,000 to 20,000 VCE frames from 15 patients & - & TP rate - 96.76\%, FP rate - 3.87\% \\
\hline
A. Murthy et al. \cite{murthi2016automatic} & 2016 & GLCM using ANN, KNN & - & 97.25\% & Sensitivity - 93.25\%, Specificity - 97.25\% \\
\hline
Chen et al. \cite{chen2017saliency} & 2017 & Saliency maps based on intensity, red-proportions & 200 VCE frames & 98.97\% & - \\
\hline
Vasilakakis et al. \cite{vasilakakis2018} & 2018 & Color descriptor, SVM, SVM-RBF & KID dataset & 81.80\% & AUC - 81.5\%, Specificity - 90.8\%, Sensitivity - 68\% \\
\hline
Hajabdollahi et al. \cite{hajabdollahi2019} & 2019 & SVM, MLP & 5 VCE bleeding frames from KID dataset, 783 frames & - & DICE score - 0.869, AUC of ROC - 0.984 \\
\hline
A. A. Kalinin et al. \cite{kalinin2020} & 2020 & CNN & 1200 VCE frames & - & IOU - 75.35, DICE Score - 84.98 \\
\hline
\end{tabular}
\end{adjustbox}
\end{table*}

\begin{table*}[htbp]
\centering
\label{table: segmentation_table2}
\scriptsize
\begin{adjustbox}{width=\textwidth,totalheight=\textheight,keepaspectratio}
\renewcommand{\arraystretch}{1.2}
\setlength{\tabcolsep}{4pt}
\begin{tabular}{|>{\RaggedRight\arraybackslash}p{2.7cm}|>{\RaggedRight\arraybackslash}p{1.5cm}|>{\RaggedRight\arraybackslash}p{2.1cm}|>{\RaggedRight\arraybackslash}p{1.7cm}|>{\RaggedRight\arraybackslash}p{1.4cm}|>{\RaggedRight\arraybackslash}p{2.8cm}|}
\hline
\textbf{Reference} & \textbf{Year} & \textbf{Technique} & \textbf{Dataset} & \textbf{Accuracy} & \textbf{Other Evaluation Metrics} \\
\hline
Kundu et al. \cite{kundu2020multiple} & 2020 & LDA, SVM-RBF & 50 VCE bleeding frames and 2588 VCE frames & 97.39\% & Precision - 87.14\%, Recall - 85.41\%, F1-Score - 86.27\% \\
\hline
Guan et al. \cite{guan2021peak} & 2021 & FCM, DPC, PGDPC, KNN & Private dataset, Kvasir-Capsule & - & IOU - 0.55 \\
\hline
H. H. Yen et al. \cite{yen2022forrest} & 2022 & Intra-rater and Inter-rater reliability score & 276 VCE frames & - & Intra-rater agreement - 0.92-0.97, Inter-rater agreement - 0.67-0.86 \\
\hline
Sheng Li et al. \cite{sheng2023seg} & 2023 & DFCA-Net & - & - & Mean IOU - 86.85\% \\
\hline
\end{tabular}
\end{adjustbox}
\end{table*}

Kundu et al. \cite{kundu2020multiple} developed a CAD method utilizing linear discriminant analysis (LDA) to separate regions of interest (ROI) and learn the characteristic patterns of diseases from pixel-labeled frames. LDA models were used to extract salient ROIs during both the training and testing phases, with intensity patterns being modeled by a suitable probability distribution. The method achieved a high accuracy, up to 96\% for bleeding video clips and 97.39\% for bleeding versus normal frames, 96.09\% for bleeding versus ulcer frames, and 97.15\% for bleeding versus tumor frames using SVM with Radial Basis Function (RBF) kernel for classification, even with a small number of pixel-based labeled frames. In their paper, Guan et al. \cite{guan2021peak} introduced a fast density peak clustering method (PGDPC) to address the challenge of preservation of local spatial information of pixels during Fuzzy c-means (FCM), reducing time complexity from $O(n^2)$ method to $O(nlogn)$ compared to Density Peak Clustering (DPC). This method enhanced super-pixel technology. It utilizes a peak-graph-based allocation strategy to enhance robustness in reconstructing spatial information for complex-shaped clusters, achieving an average IOU score of 55\% on bleeding frames from Kvasir-Capsule datsset. 

Yen et al. \cite{yen2022forrest} conducted the first quantitative analysis of bleeding ulcers from endoscopic frames, addressing differences in the interpretation of Forrest Classifications among endoscopists. They collected endoscopic frames of 276 patients with peptic ulcer bleeding and compared intra-rater and inter-rater reliability scores among endoscopists of varying expertise levels. By visually studying pixel-wise differences in Forrest Classification stages and manually delineating ulcer extent and bleeding area, they found distinct patterns for low-risk (Forrest IIC and III) in comparison to high-risk (Forrest I, IIA, or IIB) lesions, with inter-rater agreement ranging from 0.67 to 0.86. Sheng Li et al. \cite{sheng2023seg} introduced pixel-based feature extraction using dual networks, complementing each other to accurately localize bleeding points in VCE frames. These branches were based on an attention mechanism to identify each pixel as bleeding or non-bleeding, and their outputs were fused by a feature fusion module for better segmentation results. This network yielded a mean IoU of 86.85\%. 

A brief overview of the manuscripts reviewed in this sub-section is provided in the Tables \ref{table: segmentation_table1}, \ref{table: segmentation_table2}. Also, a chord chart in Fig. \ref{fig: Segmenatation_chordchart} is presented to illustrate the methodologies implemented by respective researchers in this domain. This clearly shows that SVM is the most popular ML technique for segmenting bleeding regions in VCE frames. 

\subsection{Classification and Segmentation}\label{subsec3}

G.B.Pan et al. \cite{pan2010bleeding} presented improved Euler distance for image recognition in CIE color space using the covariance matrix of VCE frames to measure color similarity, emphasizing color difference as a key feature. They constructed a pattern classifier using this improved distance, achieving a sensitivity of 92\% and specificity of 95\%, proving the algorithm's ability to identify bleeding areas correctly. Yuan et al. \cite{yuan2015} introduced a novel approach to differentiate bleeding frames from normal ones and locate bleeding areas using a color feature extraction method and K-means clustering method to derive cluster centers. They achieved 95\% accuracy and 97\% AUC with YCbCr color space with 80 clusters for SVM classification. For the localization of bleeding regions, they proposed a two-stage saliency map method based on a mixer of different color channels and visual contrast which yielded the best results using a weight of 0.8 for the first stage map and 0.2 for the second stage map, resulting in a localization precision of 95.24\%. F. Deeba et al. \cite{deeba2018} introduced a novel classifier fusion algorithm for the automatic detection of bleeding regions in VCE frames, combining two SVM classifiers based on RGB and HSV color spaces. Statistical features from first-order histograms of color channels characterize image regions. The classifiers were optimized by nested cross-validation for parameter tuning and feature selection. This approach achieved 95\% accuracy, 94\% sensitivity, and 95.3\% specificity and performed better than single classifiers.

Ghosh et al. \cite{ghosh2017cadcolor} proposed a pixel-based feature extraction method using a transform color domain, highlighting the effectiveness of the R/G (red to green) intensity ratio plane in detecting bleeding zones. They found that the G/R composite color domain yielded the best results, achieving an accuracy of 97.96\% and a specificity of 97.75\%. The following year, Ghosh et al. \cite{ghosh2018chobhs} developed a color histogram of block statistics (CHOBS) to detect bleeding in VCE videos automatically. By employing block-based local feature extraction from the RGB color space and utilizing PCA for feature reduction, CHOBS offers bleeding detection in low-feature dimensionality. This method achieved 97.85\% accuracy, 99.47\% sensitivity, and 99.15\% specificity for bleeding frame detection, and 95.75\% precision. Building on their previous study, Ghosh et al. \cite{ghosh2018cluster} introduced a new approach for automatic bleeding detection in VCE videos. Initially, block-based statistical features are derived from the transformed G/R color plane and input into K-means clustering to obtain two clusters, from which cluster-based features (CBFs) are extracted. When combined with the differential cluster features, they contribute to enhanced accuracy in supervised classification.

In their research, Liaqat et al. \cite{Liaqat2018} proposed several steps for enhancing lesion detection and classification of stomach infections, including bleeding cases using pre-processing and feature extraction from frames. Using a serial-based fusion method, they combined color, shape, and surface characteristics. Additionally, they proposed a multi-class support vector machine (MSVM) feature selection method based on PCA and correlation coefficient, which achieved high accuracy by correctly classifying up to 99\% of bleeding frames presented to it. Kundu et al. \cite{kundu2018} proposed an automatic bleeding frame and region detection method for VCE based on the intensity variation profile in the normalized RGB space. They extracted features from histograms of the variation of normalized green planes within an efficient ROI and implemented a KNN classifier, achieving high accuracy 97.86\%, sensitivity 95.2\%, and specificity 98.32\% compared to other methods. Sivakumar et al. \cite{sivakumar2019} developed an automatic obscure bleeding detection for VCE using super-pixel segmentation and Naive Bayes Classifier. They also proposed a two-stage saliency extraction method to localize bleeding mucosa from the normal regions using supervised learning methods, KNN, and SVM. The paper claimed to have improved bleeding detection and localization, but specific results were not stated.

\begin{figure}[ht]
  \centering
  \includegraphics[width=\columnwidth]{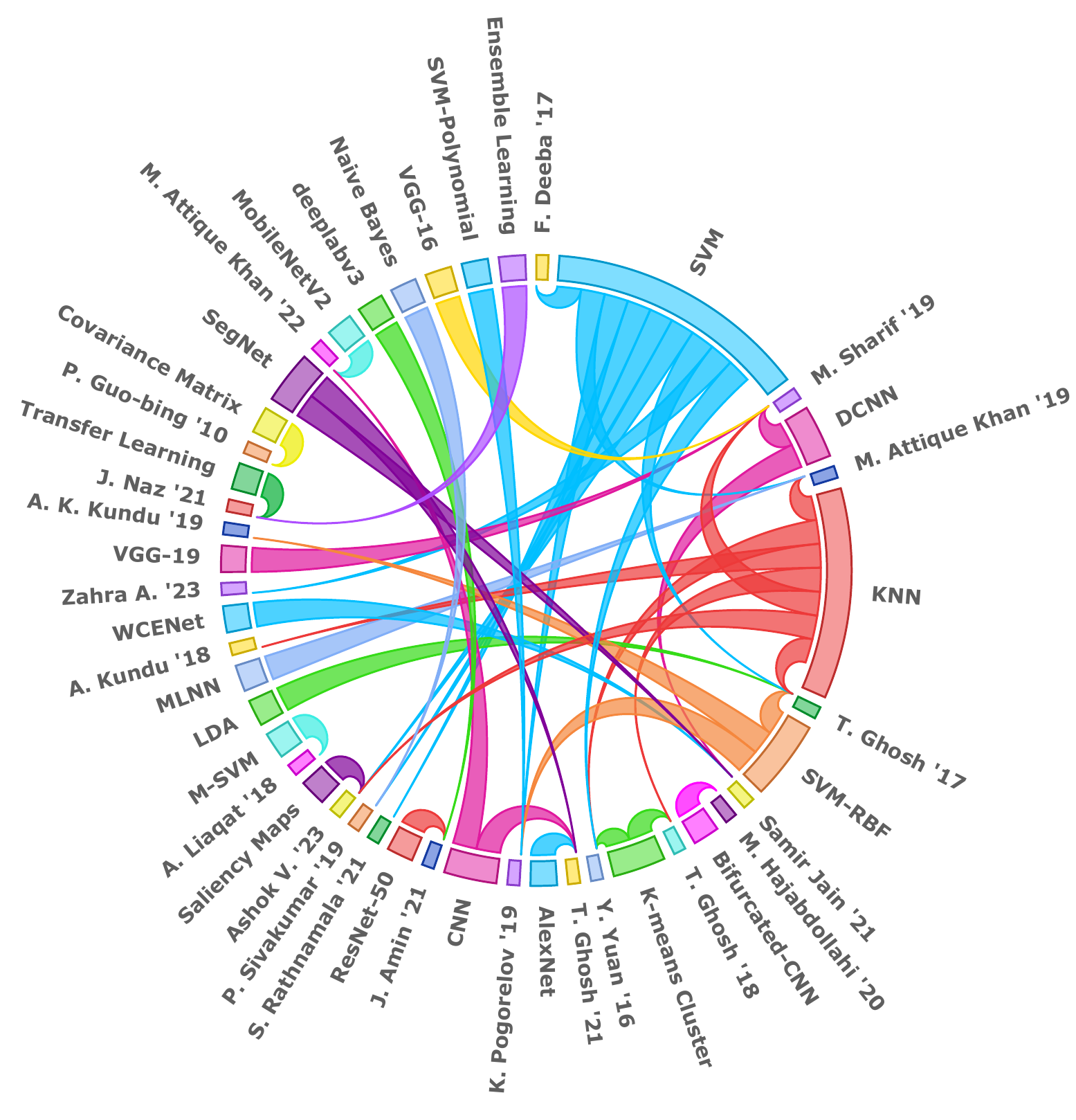}
  \caption{\RaggedRight{Authors and Methods: Chord Chart for Bleeding Frame Classification and Segmentation in VCE Images.}}
  \label{fig:Classification_Segmentation_chordchart}
\end{figure}

Pogorelov et al. \cite{Pogorelov2019} introduced an automatic bleeding detection technique in VCE videos that combines texture and color features for initial detection, followed by pixel-level segmentation of bleeding regions. They achieved 97.7\% accuracy, 97.6\% sensitivity, 95.5\% specificity, 0.978 F1-Score, and 0.898 MCC for detection of the bleeding frames. For pixel-level detection of bleeding, this algorithm achieved 97.6\% accuracy, 97.6\% sensitivity, 95.9\% specificity, 0.976 F1-Score, 0.931 MCC, and 0.997 ROC. In their research, Kundu et al. \cite{kundu2019pd} introduced a feature-based bleeding detection method that utilizes a linear separation scheme to identify pixels of interest and extract local spatial features. They fit a characteristic PDF to the resulting feature space, finding that the Rayleigh PDF model works best for bleeding detection in VCE frames. This approach provided reduced computational complexity and consistent class representation and achieved high specificity 96.59\%, sensitivity 97.55\%, and accuracy 96.77\%. Rashid et al. \cite{khan2019saliencyfs} introduced a CAD method to improve the accuracy of the detection of GI disease. Their approach involved HSI color transformation, contour segmentation, and a saliency-based method in YIQ color space. They fused frames and then features using singular value decomposition (SVD), LBP, and GLCM for the classification of bleeding, ulcer, and healthy classes. We have focused on the method's performance in the context of the detection of bleeding, and it is found that the highest accuracy achieved is 99.6\% and a sensitivity of 99.6\%, FNR of 0.4, which performs better than other established methods.

Sharif et al. \cite{Sharif2019} proposed a new technique utilizing DCNN and geometric features. They extracted bleeding regions using contrast-enhanced color features, then geometric features, and fused VGG-16 and VGG-19 CNN features based on the Euclidean Fisher Vector. This method identified 100\% of the bleeding class frames, thus performing better than previously introduced methods. M. Hajabdollahi et al. \cite{Hajabdollahi2020} addressed the computational challenges of CNN-based classification and segmentation by introducing a method that works simultaneously for both. They designed a structure with separate branches for classification and segmentation, each trained for distinct abnormalities but sharing a common part. Their method achieved a DICE score of 85\% for segmentation and 97.8\% accuracy for classification purposes of bleeding class frames. In their study, Amin et al. \cite{amin2021} proposed a novel method leveraging a deep semantic segmentation model for 3D segmentation on low-contrast VCE frames with irregularity in shape and size of lesions. Their approach has Labv3 as the backbone of ResNet-50, trained on ground masks to achieve successful pixel-wise classification. To enhance the classification accuracy, they used a pre-trained ResNet-50 model to extract deep features from global input frames, resulting in a global accuracy of 98\% and 0.96 of the mean IoU during testing. They also investigated the uncertainty and interpretability during classification through standard thresholding and Bayesian Neural Networks (BNN) techniques, achieving 100\% accuracy on privately collected datasets.

\begin{table*}[htbp]
\centering
\caption{Papers Implementing Bleeding Frame Classification and Segmentation.}
\label{table: classif_seg1}
\scriptsize
\begin{adjustbox}{width=\textwidth,totalheight=\textheight,keepaspectratio}
\renewcommand{\arraystretch}{1.2}
\setlength{\tabcolsep}{4pt}
\begin{tabular}{|>{\RaggedRight\arraybackslash}p{2.7cm}|>{\RaggedRight\arraybackslash}p{1.5cm}|>{\RaggedRight\arraybackslash}p{2.1cm}|>{\RaggedRight\arraybackslash}p{1.7cm}|>{\RaggedRight\arraybackslash}p{1.4cm}|>{\RaggedRight\arraybackslash}p{2.8cm}|}
\hline
\textbf{Reference} & \textbf{Year} & \textbf{Technique} & \textbf{Dataset} & \textbf{Accuracy} & \textbf{Other Evaluation Metrics} \\
\hline
G. B. Pan et al. \cite{pan2010bleeding} & 2010 & covariance matrix used on CIELab colorimetric system & 960 VCE frames from 10 VCE videos & - & Sensitivity-92\%, Specificity-95\% \\
\hline
Yuan et al. \cite{yuan2015} & 2016 & K-means clustering, SVM, KNN & 2400 VCE frames & 95.75\% & Sensitivity - 92\%, Specificity - 96.5\% \\
\hline
F. Deeba et al. \cite{deeba2018} & 2017 & SVM & videos from PillcamSB1 and PillcamSB2 & 97.22\% & Sensitivity - 98.13\%, Specificity - 96.52\% \\
\hline
T. Ghosh et al. \cite{ghosh2017cadcolor} & 2017 & Higher and Lower order statistical analysis, SVM-linear, SVM-RBF, KNN, LDA, Naive Bayes & 20 VCE videos, 2350 VCE frames & 98.47\% & Sensitivity 98.66\%, Specificity 99.52\% \\
\hline
T. Ghosh et al. \cite{ghosh2018chobhs} & 2018 & CHOBS, KNN, PCA & 2350 VCE frames from 32 VCE videos & 99.15\% & Sensitivity-97.85\%, Specificity-99.47\% \\
\hline
T. Ghosh et al. \cite{ghosh2018cluster} & 2018 & CBF extraction, SVM, K-means clustering & 2350 VCE frames from 32 VCE videos & 98.04\% & Precision-97.05\%, FPR-1.11, FNR-22.38, Jaccard Index-75.83\%, DICE coefficient-86.26\% \\
\hline
Liaqat et al. \cite{Liaqat2018} & 2018 & MSVM & private dataset & 98.3\% & Sensitivity-98\%, Specificity-98\%, Precision-99\%, FPR-0.054, AUC-0.99 ,FNR-1.7, G-Measure-98.49\% \\
\hline
Kundu et al. \cite{kundu2018} & 2018 & Interplane Intensity Variation Profile, KNN & 2300 VCE frames from 30 videos & 97.86\% & Sensitivity-95.2\%, Specificity-98.32\% \\
\hline
Sivakumar et al. \cite{sivakumar2019} & 2018 & Superpixel segmentation, NaiveBayes Classifier, SVM, KNN & - & - & -\\
\hline
Pogorelov et al. \cite{Pogorelov2019} & 2019 & SVM-linear, SVM-polynomial, SVM-RBF & private database & 97.7\% & Sensitivity-97.7\%, Specificity-95.5\% \\
\hline
Kundu et al. \cite{kundu2019pd} & 2019 & SVM-RBF & 2350 VCE frames from 2350 VCE videos & 96.77\% & Sensitivity- 97.55\% ,Specificity- 96.59\% \\
\hline
Rashid et al. \cite{khan2019saliencyfs} & 2019 & SVD, LBP, GLCM, MLNN, SVM, KNN & 9000 VCE frames & 87.09\% & - \\
\hline
\end{tabular}
\end{adjustbox}
\end{table*}

\begin{table*}[htbp]
\centering
\label{table: classif_seg2}
\scriptsize
\begin{adjustbox}{width=\textwidth,totalheight=\textheight,keepaspectratio}
\renewcommand{\arraystretch}{1.2}
\setlength{\tabcolsep}{4pt}
\begin{tabular}{|>{\RaggedRight\arraybackslash}p{2.7cm}|>{\RaggedRight\arraybackslash}p{1.5cm}|>{\RaggedRight\arraybackslash}p{2.1cm}|>{\RaggedRight\arraybackslash}p{1.7cm}|>{\RaggedRight\arraybackslash}p{1.4cm}|>{\RaggedRight\arraybackslash}p{2.8cm}|}
\hline
\textbf{Reference} & \textbf{Year} & \textbf{Technique} & \textbf{Dataset} & \textbf{Accuracy} & \textbf{Other Evaluation Metrics} \\
\hline
Sharif et al. \cite{Sharif2019} & 2019 & KNN, DCNN & private dataset with 5500 VCE frames & 99.42\% & Precision rate-99.51\% \\
\hline
Hajabdollahi et al. \cite{Hajabdollahi2020} & 2020 & Bifurcated CNN & KID and private bleeding dataset & 96\% & Sensitivity-92.4\%, Specificity-97.8\% \\
\hline
Amin et al. \cite{amin2021} & 2021 & BNN, ResNet, deeplabv3 & 30 VCE videos, each containing 500 frames, CVC–Clinic DB, The Nerthus dataset, Kvasir-Capsule for Segmentation and Classification & 98\% & - \\
\hline
Ghosh et al. \cite{ghosh2021deeptransfer} & 2021 & CNN, AlexNet, SegNet & KID dataset & 94.42\% & - \\
\hline
Jain et al. \cite{jain2021} & 2021 & VCENet, SegNet, DCNN & KID dataset & 98\% & ROC-99\%, freq. weighted IOU-81\%, avg DICE score-56\% \\
\hline
S. Rathnamala et al. \cite{rathnamla2021} & 2021 & SVM-binary, GMM & Dataset1 contains 3294 VCE frames, and Dataset2 contains 100 VCE frames & 99.61\% & Specificity- 99.87\%, Sensitivity- 98.49\%, Precision- 0.99, DICE score 0.99, Jaccard Index- 0.98 \\
\hline
Naz et al. \cite{naz2021segmentation} & 2021 & TL, LBP, Ensemble Learning Classifier & Private and Kvasir dataset& 99.8\% & Jackard Index- 96.58\%, FNR-6.61\\
\hline
Khan et al. \cite{khan2022gestronet} & 2022 & CNN & Kvasir 1, Kvasir 2, CUI Wah & 99.61\% & - \\
\hline
Zahara Amiri et al. \cite{amiri2023combining} & 2023 & TL, LBP, HOG, ResNet, SVM & Kvasir-Capsule, GIANA 2017, RLE, KID & 99.6\% & FNR-0.008, FPR-0.003\%, Precision-0.99, Recall-0.992, F-Measure-0.991, MCC-0.988\%\\
\hline
Ashok et al. \cite{Ashok2023} & 2023 & Saliency maps based on optical contrast and color channel mixers, SVM, KNN & - & 97.75\% &-\\
\hline
\end{tabular}
\end{adjustbox}
\end{table*}

%
%



Ghosh et al. \cite{ghosh2021deeptransfer} introduced a CAD tool for the automatic analysis of VCE frames, focusing on the detection of bleeding in the small intestine. They used a pre-trained AlexNet to train a CNN for the detection of bleeding and non-bleeding VCE frames and incorporated SegNet for the delineation of bleeding regions in frames. This model was evaluated on two publicly available datasets, where it achieved a global accuracy of 94.42\% and a weighted IoU of 90.69\% for detecting bleeding zones. Jain et al. \cite{jain2021} introduced WCENet, a CNN-based model, for the detection and localization of GI anomalies in VCE frames. Initially, an attention-based CNN classifies the frames into four categories, and then, for anomaly localization, a Grad-CAM++ and custom SegNet fusion is employed. 
The WCENet classifier obtained 98\% accuracy, and the segmentation model achieved a frequency-weighted IoU of 81\% and an average dice score of 56\% on the KID dataset for the bleeding class. S. Rathnamala et al. \cite{rathnamla2021} proposed an automatic bleeding detection system for VCE frames utilizing features from Gaussian mixture model (GMM) superpixels and binary SVM classification followed by segmentation based on deltaE color differences. They achieved higher classification and segmentation accuracy of 99.88\% with less computational time.

Naz et al. \cite{naz2021segmentation} proposed two procedures for bleeding region segmentation and GI abnormality classification using DL. They applied a hybrid approach for bleeding segmentation, a threshold to RGB channels, and pixel-wise comparison to generate masks. For the classification of bleeding, texture features using LBP and deep features were extracted by applying TL and the best ones were input into the Ensemble Classifier. This method achieved the DICE score of 93.39\% and 96.58\% in Jack-index for segmentation and accuracy of 99.7\% on the private dataset and 86.6\% on the Kvasir-Capsule dataset for classification. Khan et al. \cite{khan2022gestronet} developed a framework using deep saliency maps and Bayesian optimal DL. They improved image quality through contrast enhancement and implemented pre-trained, fine-tuned MobileNet-V2, initializing hyperparameter using Bayesian Optimization. A hybrid whale optimization algorithm was introduced to select optimal features, followed by classification using an extreme learning classifier. The framework achieved 99.8\% accuracy in identifying the bleeding class, showcasing its effectiveness. Zahara Amiri et al. \cite{amiri2023combining} proposed a CAD method for extraction of more distinct image areas in the background as ROI based on color and texture characteristics. They utilized a ResNet-50 pre-trained model to extract deep features and proposed a new method for segmentation for VCE frames using Expectation Maximization (EM) initialization to improve accuracy. A subset of the features generated by this method was selected using the approach of minimum redundancy and maximum relevance and was then classified using SVM, achieving an accuracy of 99.8\%, precision of 99.2\%, recall of 99.6\% and an MCC of 99.3\%. Ashok et al. \cite{Ashok2023} introduced a novel color extraction method to discriminate between bleeding and non-bleeding frames. Their approach involved two steps: analyzing color information in bleeding frames and deriving cluster centers from pixel-based representation, which returned pixels as words. Saliency maps for localization of bleeding region were generated by applying distinct color channel mixers and optical contrast with appropriate thresholding and fusion techniques. This framework achieved 95.75\% accuracy for classification purposes. 

Fig. \ref{fig:Classification_Segmentation_chordchart} is the chord chart depicting the authors in this sub-section and the corresponding techniques implemented by them. A similar observation is noted here that SVM is the most used technique used here, followed by KNN. It should also be noted that more deep-learning techniques have been implemented here than in the previous sub-section. Tables \ref{table: classif_seg1} and \ref{table: classif_seg2} provides details of the papers reviewed in this section, such as the techniques implemented. datasets used and the results achieved.

\subsection{Detection}\label{sec2}

\begin{figure}[ht]
  \centering
  \includegraphics[width=\columnwidth]{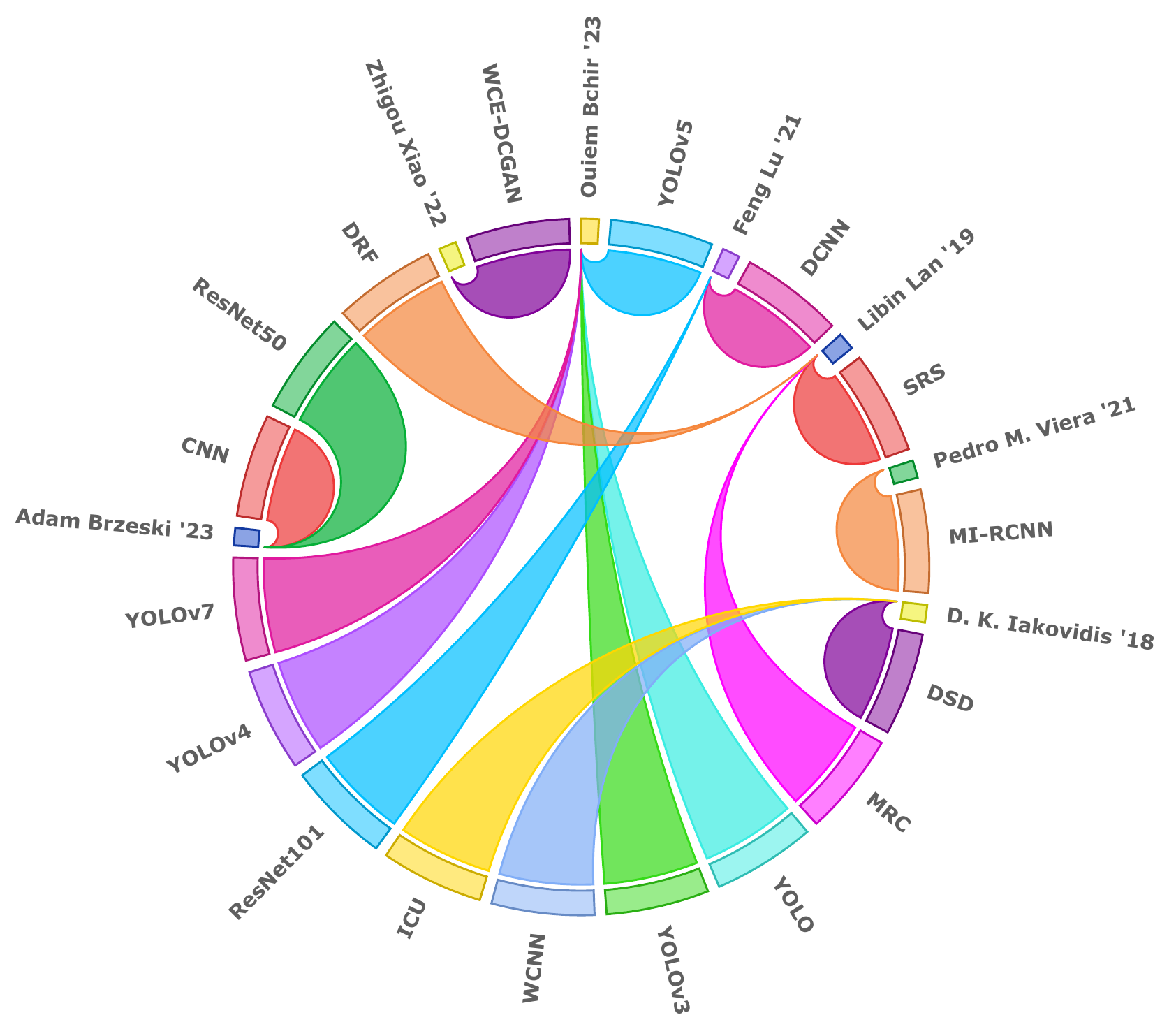}
  \caption{\RaggedRight{Chord Chart of Authors and Techniques in Bleeding Frame Detection for VCE images.}}
  \label{fig:Detection_chordchart}
\end{figure}

Iakovidis et al. \cite{iakovidis2018detecting} introduced a weakly-supervised methodology employing a CNN for the automatic detection and localization of GI anomalies, which included bleeding among other conditions. Utilizing image-level annotations, the model classifies frames as normal or abnormal detects salient features via a Deep Saliency Detection algorithm and localizes anomalies using an Iterative Cluster Unification algorithm. While specific results for bleeding-only detection were not isolated, the datasets employed in the study did contain bleeding frames. The approach, tested on datasets with various GI anomalies, demonstrated an AUC of over 80\%, with the highest AUC for anomaly detection at 96\% in gastroscopy frames and 88\% for anomaly localization in VCE frames. In addressing the challenge of identifying abnormal patterns in VCE frames, Libin Lan et al. \cite{lan2019deep} propose a DL approach utilizing a CascadeProposal network. This network combines a region proposal rejection module for high-recall region proposals and a detection module for abnormal pattern recognition, improved by a multiregional combination (MRC) method, salient region segmentation (SRS), and dense region fusion (DRF) for enhanced localization. Additionally, a negative category and TL strategies are integrated to refine model performance. The methodology achieved a mean average precision (mAP) of 72.3\% specifically for detecting bleeding abnormalities within a dataset of over 7,000 annotated frames. 

Feng Lu et al. \cite{FengLuDetection} developed an edge analytic framework for detecting and localizing bleeding lesions in VCE images using a deep multi-scale feature fusion network. The approach integrates a multi-scale regional proposal network with a top-down feature fusion, enhancing the resolution and precision of detection. Validation against a clinical trial dataset showed a recall rate and F1 score of 31.69\% and 22.12\%, respectively, with a prediction accuracy of 98.9\% and a 4.88\% Average Precision (AP). Vieira et al. \cite{VieiraDetection2021} proposed a multi-pathology detection system for VCE frames, utilizing Mask Improved RCNN (MI-RCNN) with mask subnet and training strategy based on second momentum. The paper focused on the detection and localization of various pathologies, such as bleeding, angioectasias, polyps, and inflammatory lesions in VCE frames. Tested on the KID public database, the system demonstrated performance metrics with mAP, mAP50, mAP75, and F1-Score of 40.35, 59.42, 43.01, and 60.07, respectively. Xiao et al. \cite{XiaoDetection2022} developed a WCE-Deep Convolutional GAN (DCGAN) network to tackle the challenge of small dataset sizes and uneven distribution in VCE image datasets, focusing on pathologies such as Ulcer, Polyp, Blood—Fresh, and Erosion. This network, aimed at enhancing data diversity and volume, was tested across object detection frameworks, including Single-Shot Detector (SSD), YOLOv5, and YOLOv4. Specifically for Blood—Fresh pathology detection, the augmented dataset yielded an accuracy of 99.2\% with YOLOv5 and 99.98\% with SSD, indicating the network's efficacy in generating diverse, generalizable data for improved object detection performance in VCE frames.

\begin{table*}[htbp]
\centering
\caption{Overview of Papers with Bleeding Region Detection.}
\label{table: detection}
\scriptsize
\begin{adjustbox}{width=\textwidth,totalheight=\textheight,keepaspectratio}
\renewcommand{\arraystretch}{1.2}
\setlength{\tabcolsep}{4pt}
\begin{tabular}{|>{\RaggedRight\arraybackslash}p{2.7cm}|>{\RaggedRight\arraybackslash}p{1.5cm}|>{\RaggedRight\arraybackslash}p{2.1cm}|>{\RaggedRight\arraybackslash}p{1.7cm}|>{\RaggedRight\arraybackslash}p{1.4cm}|>{\RaggedRight\arraybackslash}p{2.8cm}|}
\hline
\textbf{Reference} & \textbf{Year} & \textbf{Technique} & \textbf{Dataset} & \textbf{Accuracy} & \textbf{Other Evaluation Metrics} \\
\hline
Iakovidis et al. \cite{iakovidis2018detecting} & 2018 & CNN & Kvasir 1, Kvasir 2, CUI Wah & - & AUC over 80\% \\
\hline
Lan et al. \cite{lan2019deep} & 2019 & CNN, TL & 7000 frames & - & Mean average precision (mAP) - 72.3\% \\
\hline
Feng Lu et al. \cite{FengLuDetection} & 2021 & CNN & 20,000 VCE frames & 98.9\% & AP - 4.8\%, Recall - 31.69\%, 22.12\% \\
\hline
Vieira et al. \cite{VieiraDetection2021} & 2021 & Mask Improved-RCNN & KID & - & mAP - 40.35, mAP50 - 59.42, mAP75 - 43.01, F1-Score - 60.07\\
\hline
Xiao et al. \cite{XiaoDetection2022} & 2022 & DCGAN, SSD, YOLO & Kvasir-Capsule  & 99.2\% with YOLOv5 and 99.98\% with SSD & - \\
\hline
Bchir et al. \cite{bchir2023deep} & 2023 & YOLO & Kvasir-Capsule & - & mAP - 0.883, IoU - 0.8 \\
\hline
Brzeski et al. \cite{brzeski2023visual} & 2023 & CNN & ERS & - & AUC of 0.963 \\
\hline
\end{tabular}
\end{adjustbox}
\end{table*}

Bchir et al. \cite{bchir2023deep} designed a pattern recognition system using YOLO DL models to detect MBS. The study employed YOLOv3, YOLOv4, YOLOv5, and YOLOv7 to recognize MBS, with YOLOv7 achieving the highest performance metrics, specifically a mAP of 0.86 and an IoU of 0.8. The effectiveness of YOLOv7 was further confirmed by data augmentation, which led to a mAP of 0.883 for bleeding spot detection. The study by Adam Brzeski et al. \cite{brzeski2023visual} explored bleeding detection in endoscopic videos through binary classification using a CNN enhanced with high-level visual features. The method introduces domain-specific feature descriptors that output activation maps, augmenting the input data to the CNNs. For bleeding detection, the feature-extended models showed promising performance in comparison to the baseline models. Specifically, the Resnet50 architecture augmented with these features achieved an AUC of 0.963.

In the chord chart provided in Fig. \ref{fig:Detection_chordchart}, it can be understood that the YOLO algorithm has been implemented by various authors in this domain. It should be noted that novel CNN algorithms have also been introduced for the purpose of the detection of bleeding regions in VCE frames. A review of the papers in this section has been provided in the Table \ref{table: detection}.

\section{Conclusion and Future Research Directions}\label{sec6}

Our analysis of 113 papers from 2008 to 2023 demonstrates the evolving landscape of research in this domain. However, despite significant advancements, several challenges and opportunities for future research emerge:

\begin{enumerate}

\item \textbf{Model Generalization}: Future studies should focus on developing generalized models that can robustly detect GI bleeding across diverse datasets, addressing variations in patient demographics and pathological presentations.

\item \textbf{Clinical Implementation}: Although ML models have shown promise in research settings, their integration into clinical practice remains limited. Efforts should be directed towards validating these models in clinical trials and understanding the barriers to their practical deployment.

\item \textbf{Data Diversity and Availability}: There is a need for larger, more diverse datasets that include a wide range of bleeding instances and patient backgrounds to train more robust and universally applicable models.

\item \textbf{Interpretable ML Models}: The black-box nature of many ML models hampers their acceptance in clinical practice. Future research should strive to improve the transparency and interpretability of these models, allowing healthcare professionals to understand and trust the predictions made by ML algorithms.

\item \textbf{Cross-disciplinary Collaboration}: Encouraging collaborations between computer scientists, biomedical engineers, and clinicians can lead to the development of more effective and clinically relevant ML solutions.
\end{enumerate}
This review has provided a comprehensive examination of ML techniques for GI bleeding detection in VCE frames, highlighting significant advancements, evaluating the effectiveness of various methodologies, and discussing their challenges and future prospects. The insights drawn from this comprehensive review aim to not only guide future researchers but also contribute foundational knowledge toward the development of more efficient diagnostic methods. By addressing the identified challenges and exploring suggested future directions, the research community can enhance diagnostic methods, ultimately advancing GI healthcare.

\section*{Acknowledgment and Declarations}

\label{sec:decalarations}
\begin{itemize}
\item Funding: Not applicable (NA) 
\item Competing Interest: The authors declare no personal, academic and financial conflicts of interests associated with this study.
 \item Availability of meta-data: NA
 \item Ethical Approval: NA
 \item Consent to participate: NA 
 \item Consent to publish: NA
\end{itemize}

The authors declare no personal, academic, or financial conflicts of interest associated with this work. No funding was received for this work.

\bibliographystyle{IEEEtran}
\bibliography{template/citation} %

\end{document}